\documentclass[10pt,journal,compsoc]{IEEEtran}

\newcommand\rebuttal{false}

\usepackage{amsmath} % assumes amsmath package installed
\usepackage{amssymb}  % assumes amsmath package installed
\usepackage{latexsym}
\usepackage{url}
\usepackage{relsize}
\usepackage{multirow}
\usepackage{afterpage}
\usepackage{bm}
\usepackage{ifthen}
\usepackage{color}
\usepackage{soul}
\usepackage{enumerate}
\usepackage[pdftex]{graphicx}
\usepackage[multiple]{footmisc}
\graphicspath{{./figures/}}

\def\beqn#1\eeqn{\begin{eqnarray}#1\end{eqnarray}}
\def\beq#1\eeq{\begin{equation}#1\end{equation}}
\def\bea#1\eea{\begin{align}#1\end{align}}
\def\beg#1\eeg{\begin{gather}#1\end{gather}}
\def\beqs#1\eeqs{\begin{equation*}#1\end{equation*}}
\def\beas#1\eeas{\begin{align*}#1\end{align*}}
\def\begs#1\eegs{\begin{gather*}#1\end{gather*}}
\def\bpm#1\epm{\begin{pmatrix}#1\end{pmatrix}}

\renewcommand{\vec}[1]{\mathbf{#1}} %looks better
\newcommand{\uvec}[1]{\hat{\vec{#1}}}
\newcommand{\vecmb}[1]{\boldsymbol{#1}} %handles better zero vectors

\newcommand{\diag}{\mathop{\mathrm{diag}}}

\newcommand{\sgn}{\mathop{\mathrm{sign}}}

\newcommand{\avg}{\mathop{\mathrm{avg}}}

\newcommand{\erf}{\mathop{\mathrm{erf}}}

\newcommand{\R}{\mathbb{R}}
\newcommand{\defeq}{\triangleq}

\newcommand{\junk}[1]{}

\newcommand{\quat}[1]{\ensuremath{\mathring{\mathbf{#1}}}}

\newcommand{\fixerror}[2]{\ifthenelse{\boolean{\rebuttal}}{\textcolor{red}{#1 \st{#2}}}{{#1}}}

\ifCLASSOPTIONcompsoc
  \usepackage[nocompress]{cite}
\else
  \usepackage{cite}
\fi

\ifCLASSINFOpdf
\else
\fi

\begin{document}
\title{Curved patch mapping and tracking for irregular terrain modeling: \\
Application to bipedal robot foot placement}
\author{Dimitrios~Kanoulas, Nikos G. Tsagarakis, and Marsette~Vona% <-this % stops a space
  \IEEEcompsocitemizethanks{
    \IEEEcompsocthanksitem Dimitrios Kanoulas and Nikos G. Tsagarakis are with the Humanoids and Human-Centered Mechatronics Lab, Instituto Italiano di Technologia, Via Morego 30, 16163, Genova, Italy.\protect\\
    E-mail: {\{Dimitrios.Kanoulas,Nikos.Tsagarakis\}@iit.it}
    \IEEEcompsocthanksitem Marsette Vona is with the NASA Jet Propulsion Laboratory (JPL), Pasadena, CA, 91109, USA.\protect\\
    E-mail: {vona@jpl.nasa.gov}
    \IEEEcompsocthanksitem Dimitrios Kanoulas and Marsette Vona were at Northeastern University in Boston, MA, when the bulk of this work was performed, with support from the National Science Foundation under Grant No. 1149235.
}}

\IEEEtitleabstractindextext{%
\begin{abstract}
Legged robots need to make contact with irregular surfaces, when operating in unstructured natural terrains. Representing and perceiving these areas to reason about potential contact between a robot and its surrounding environment, is still largely an open problem. This paper introduces a new framework to model and map local rough terrain surfaces, for tasks such as bipedal robot foot placement. The system operates in real-time, on data from an RGB-D and an IMU sensor. We introduce a set of parametrized patch models and an algorithm to fit them in the environment. Potential contacts are identified as bounded curved patches of approximately the same size as the robot's foot sole. This includes sparse seed point sampling, point cloud neighborhood search, and patch fitting and validation. We also present a mapping and tracking system, where patches are maintained in a local spatial map around the robot as it moves. A bio-inspired sampling algorithm is introduced for finding salient contacts. We include a dense volumetric fusion layer for spatiotemporally tracking, using multiple depth data to reconstruct a local point cloud. We present experimental results on a mini-biped robot that performs foot placements on rocks, implementing a 3D foothold perception system, that uses the developed patch mapping and tracking framework.
\end{abstract}

}

\maketitle

\section{Introduction} \label{Sec:intro}
Over the past decades, bipedal robots have gained the capability
to operate and locomote in well-structured environments.  Still,
their reliability is not guaranteed for real-world unstructured
trails, where the legged robots need to contact uneven surfaces,
under significant uncertainty.  The profound impact that the
Fukushima Daiichi nuclear disaster had on robotics in 2011,
highlighted the need for robots to replace humans in hazardous
tasks, such as climbing and searching rubble piles after a
disaster~\cite{KS09} or operating in human-traversable rough
terrain.  The use of 3D visual perception can be crucial for
completing tasks that require bipedal robots locomotion in uneven
3D terrain.

\begin{figure*}
\includegraphics[width=\textwidth]{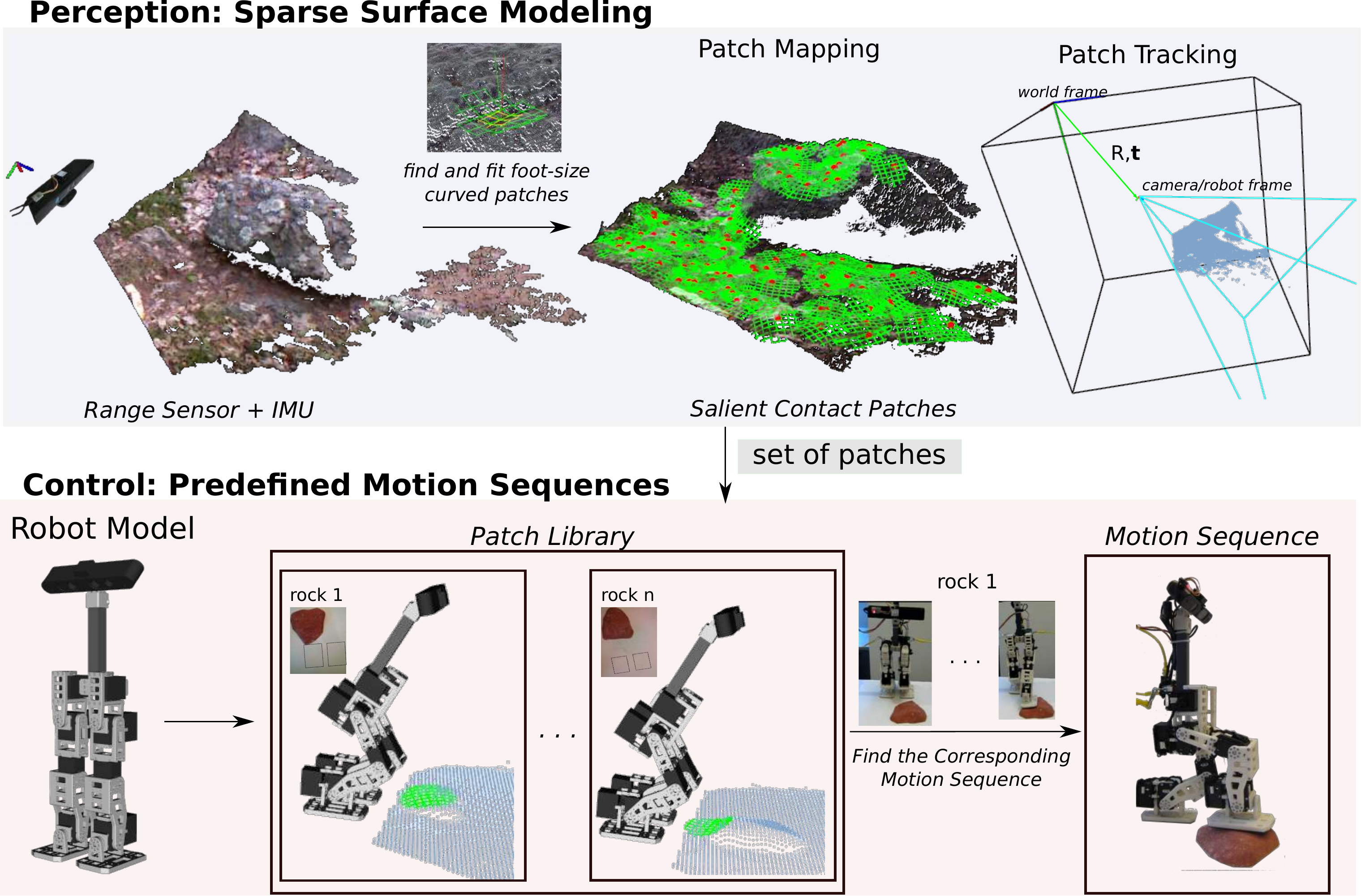}
\caption{Overview of the total curved patch mapping and tracking system, which splits in two parts: perception (top) and motion control (bottom).\label{Fig:overview}}
\end{figure*}

Most of the current work in bipedal control, planning, and
perception has focused on flat surfaces
locomotion~\cite{CLCKHK05,MCKK2005,NCK12}, where the environment
is mostly known or well structured.  Only recently, the work has
been extended to outdoors rough terrain environments, where
bipedal robots are gaining capability mostly without using 3D
perception.  The uncertainty is then tolerated either by
low-level feedback control for blindly landing
footholds~\cite{RBNP08,Wiedebach2016} or by using tactile or
proprioceptive sensing.  Thus, the problem of 3D perception for
unmodeled sparse curved terrain is still largely open.  In this
work, we focus primarily on the perception part of bipedal
locomotion for rough terrain foothold placement, using both
exteroceptive and proprioceptive sensing, where non-point feet
make contact with patches of the terrain.  Notice, that
quadrupedal/hexapedal locomotion, uses point-like contact
modeling, since the feet are usually spherical. That differs from
bipeds, which may need to have multiple (continuous or discrete)
contact points with the environment.

This paper, introduces a novel method to detect and model sparse
areas for potential contact, using curved patches that are
spatially mapped in real-time around the robot.  The system is
divided into two parts (Fig.~\ref{Fig:overview}).  In the perception
part, the sparse surface modeling takes place.  For this, we
introduce: (1) a model to represent local contact surface areas of
the same scale as surfaces on robot's foot sole, (2) a fast method
to find these areas, considering uncertainty in the data, and (3)
a localization and mapping system of the detected contact areas,
while the robot is moving.  The second part, implements a simple
control system, that is based on predefined motion sequences,
which are matched to the detected patches around the robot.  Our
planning and control are basic, as we are focused mostly on
perception; many other researchers have worked on motion for this
type of task~\cite{MLB13,FMDWAMT15}.

The main contributions of this paper are as follows.  First, a new sparse contact surface representation is introduced, using a set of bounded curved patches that can model regions both in the environment and on the robot.  A fast algorithm is developed, to fit these patches to 3D surface point cloud data, with quantified uncertainty both in the input points and in the output patch.  Moreover, a residual, coverage, and curvature validation method takes place, to evaluate the quality and fidelity of the fitted patches.  Last, a bio-inspired method is introduced, to sample salient contact patches in the environment, statistically similar to those selected by humans, when hiking in rough rocky terrain.  The developed algorithms are part of a real-time mapping and tracking system, for patches around the robot, which are potentially good for contact.  The framework is experimentally validated on the newly developed mini-biped robot, called Rapid Prototyped Biped (RPBP), for foot placement contacts on rocky surfaces.

Some earlier results of this work were presented in prior
conference papers~\cite{VK11,KV13,KV14}.  This paper, is the
first presentation of the work as a whole, including also the
patch tracking system (Sec.~\ref{Sec:tracking}) and the hardware
experiments with a bipedal robot (Sec.~\ref{Sec:exp}).

Next, we cover related work (Sec.~\ref{Sec:rw}) followed by a
description of the input perceptual data (Sec.~\ref{Sec:input}) and
the sparse surface modeling using bounded curved contact patches
(Sec.~\ref{Sec:patches}).  We then, introduce the patch mapping
(Sec.~\ref{Sec:mapping}) and tracking (Sec.~\ref{Sec:tracking}) system.
Finally, we present experimental results (Sec.~\ref{Sec:exp}),
including the task of foot placement on rocks, using a newly
developed mini-biped robot, with a range camera and an IMU.  The
perception algorithms are developed and released as the Surface
Patch Library (SPL)~\cite{SPL14}.

\subsection{Related Work}\label{Sec:rw}
Bipedal locomotion advanced a lot during the DARPA Robotics
Challenge (DRC) in 2015~\cite{DRC-what-happened}.  Prior to the
DRC, only few bipedal robots were using on-line foot placement
methods, based on visual/range sensing, especially for rough or
uneven terrain.

Most of the early works, focused on flat indoor environments and
usually were not real-time systems.  RGB data were originally
used in~\cite{CLCKHK05,MCKK2005} for 2D path planning, applied on
the ASIMO humanoid robot.  Similarly, in~\cite{GHB11,HDLB12},
range sensing was used for a 2D graph-based footstep planner,
implemented on the mini-bipedal robot NAO.  Real-time obstacle
re-planning methods, for dynamically changing flat 2D terrains,
were also introduced in~\cite{KB16}, tested on the NAO and HRP-4
robots.  Climbing stairs/obstacles, was also one of the primary
use-cases for bipedal systems.  The full-size humanoid robot
HRP-2, used stereo data for extracting planar surfaces to
navigate on flat terrains and climb horizontal
stairs~\cite{OII03,OOHI05}.  The humanoid robot QRIO, achieved
climbing on indoor sloping and elevated terrain, using range
data~\cite{GFF08}.  Similar concepts were studied
in~\cite{CTSNKK09,NCK12}, using 3D laser sensing.  3D perception
for footstep planning was also used in~\cite{MLB13,UNOI2014}, to
allow a NAO robot navigate on surfaces with obstacles, either by
avoiding or stepping on them.    A multi-contact method was
introduced in simulation for HRP-2~\cite{BVKEK13}, using point
cloud data for planar contact reasoning.
In~\cite{RGMSHS13,KMK13}, point cloud data were used to avoid
harsh foot impacts on a simulated HRP-2, using a
KinectFusion-based~\cite{NIHMKDKSHF11} mapping system.

Exteroceptive-based perception (vision and range sensing) methods
were developed during the DRC 2015, for footstep planning on flat
uneven outdoors terrain.  In~\cite{DT2014,FMDWAMT15}, a real-time
optimization-based state estimation system was introduced for the
ATLAS humanoids robot, where stereo fusion was used for planning
locomotion on rough planar terrain, by extracting planar contact
surfaces for navigation.  In a different direction, a graph-based
footstep planner was used
in~\cite{Kohlbrecher14,SKSC2016,KOB2016}, using LiDAR/range
environment data.  Human supervision~\cite{SKCS14} was mostly
necessary for adaptations in the foothold poses.  Most of the
aforementioned methods, localize footholds with full-surface
contact between a flat surface and the robot foot sole.  This is
done by segmenting planes in the environment.  In this work, we
differ by modeling and localizing sparse curved patches for
contact (potentially partial) between the legged robot and the
environment.  Notice, that Supraped, as introduced
in~\cite{KC14}, assumes only a single foot contact point with the
terrain and was tested only in simulation, without the use of
vision.

On-line footstep planning for quadrupeds and hexapods has also a
significant history in legged robot locomotion.  Even though they
differ from bipedal research, since usually multi-legged
platforms with spherical feet assume point-like contacts with the
environment, compared to bipeds, where area-contact surfaces are
required.  Originally, many multi-legged robots used visual
odometry and perception for obstacle avoidance and traversability
analysis~\cite{WMBHRR10,SHG12}, but not for planning 3D foot
placement.  Over the past decade, on-line perception systems for
foothold detection were also introduced.  For instance, dense
surface models were used in~\cite{PMPKRB09,KKN09,KBPMS10} for
LittleDog locomotion, while a local decision surface was used
in~\cite{BPP10} for a hexapodal robot.  Moreover, perception was
combined with supervised learning for footstep
placement~\cite{KBPMS10}, while visual SLAM systems, such as the
Parallel Tracking and Mapping (PTAM), were used with elevation
maps during locomotion~\cite{BS12}.  These systems were not
usually considering uncertainty in the foothold contact areas,
and in some cases were not real-time.  More recently, three
impressive systems were introduced
in~\cite{WFDHKS15,MHWCS15}, and~\cite{KB2017}, where the
quadruped robots StarlETH and HyQ, and the mobile/legged hybrid
quadruped robot Momaro, correspondingly, used real-time updated
elevation maps for footstep planning.

Both exteroceptive and proprioceptive perception were used in
legged robots to detect and localize footholds, either from a
distance or during contact.  Systems such as~\cite{CPP99} used
only proprioception, while others such as~\cite{MHB12,BVKEK13},
that were described above, used range sensing to drive high-level
actions.  Data fusion was used recently to deal with sensor noise
and uncertainty, as in~\cite{WFDHKS15}.  3D range uncertainty
plays an important role for foothold placement evaluation.
Various methods were used in the literature to represent
uncertainty, including both Gaussian~\cite{FAF86} and
non-Gaussian~\cite{MS87} noise modeling for 3D stereo
measurements.  Recently, an uncertainty model for the Kinect
sensor was introduced in~\cite{KE12}, while a mixture of
Gaussians was used in~\cite{DVX13}.  In this work, we will
consider 3D Gaussian noise for the input point cloud data, since
it represents fairly good the point cloud uncertainty that is
introduced from our range sensors (more details in
Sec.~\ref{Sec:input}).

A main contribution of this work is the introduction of a new
environment model, for localizing potential contact surfaces for
bipedal robots, using uncertain range sensing.  Our approach is
distinct from range image
segmentation~\cite{HBJFBGBEFF96,PBJB98}, in which the environment
is partitioned into non-overlapping regions---our patches can
overlap, but may enable better modeling without having to
consider foot contact against more than a region at a time.  In
legged locomotion, usually dense approaches are used, such as
grid or elevation maps~\cite{MLB13}.  In these approaches, the
planner evaluates the whole terrain to find areas for contact,
which is often computationally expensive.  There are also sparse
approaches, but usually planar patches are searched
for~\cite{FMDWAMT15}, without considering uncertainty.

Another main result in our work, is an algorithm to fit bounded
curved patches to noisy point cloud data.  Even though plane
fitting was studied extensively in the past, including
uncertainty~\cite{WTHNY01,Kanatani05} for heteroskedastic range
data~\cite{PVB09}, quadrics can be a better option for natural
rough terrain local contact modeling.  Quadric fitting was
studied in~\cite{Petitjean02,DNC07}, but we are the first to
combine it with quantified uncertainty, using geometrically
meaningful parametrization (vs algebraic parametrization) and
bounded patches.

To our knowledge, the idea we present in this paper, of mapping
sparse curved patches around a biped for locomotion, was not
studied before.  Some early papers, used SLAM algorithms to map
flat surfaces~\cite{MDR04,FMDWAMT15}, where either the whole
environment is segmented into planes or a local area around the
robot.  In our approach, we map a sparse set of patches, that
have approximately the size of potential contact surfaces on the
robot.  For the mapping, we customized a version of the Moving
Volume KinectFusion algorithm~\cite{RV12}, that builds a dense 3D
volumetric map of the environment around the robot, while
simultaneously it is tracking the pose of the range sensor in it.

\section{Input data: range and IMU sensing}\label{Sec:input}
Both \textit{exteroceptive} and \textit{proprioceptive}
perception are essential for foothold placement, during legged
robot locomotion in a real world environment.  From the various
sensors that have been introduced, e.g.~in~\cite{Everett95,SNS11}, we use range sensing to detect
upcoming 3D contacts and inertial measurement unit (IMU) sensing
to acquire the robot's orientation relative to gravity.  Next, we
cover the data acquisition, including range data noise analysis
and the IMU calibration process.
%\QUERY[6]

\begin{figure*}
\includegraphics[width=\textwidth]{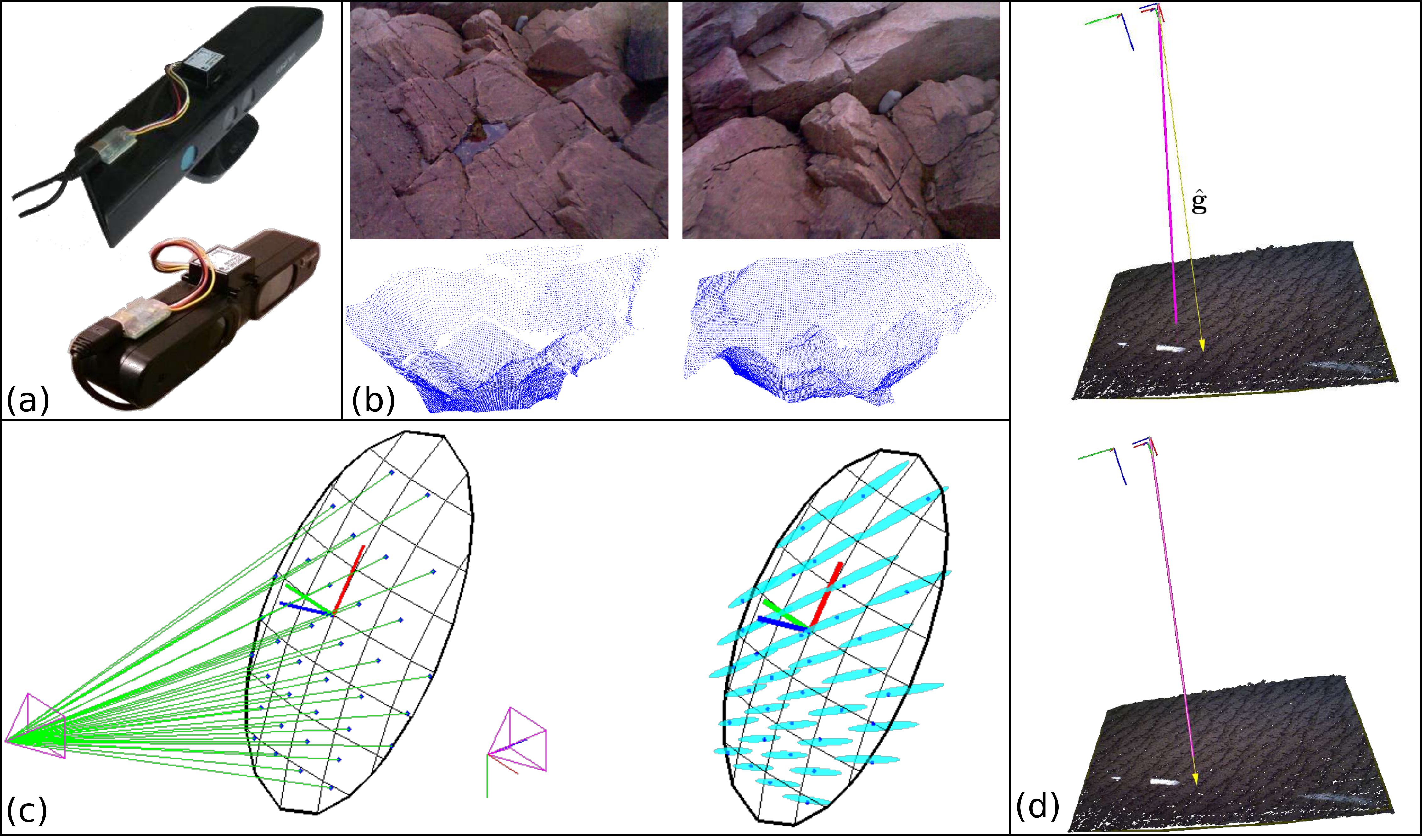}
\caption{{(a)} A Microsoft Kinect (top) and a Primesense Carmine 1.09 (bottom) RGB-D sensor, embedded with a CH Robotics UM6 9 -DoF IMU sensor; (b) a {$640~\times~480$} dense point cloud from a Microsoft Kinect RGB-D camera;  (c) the stereo error modeling, visualizing the $95\%$ probability error ellipsoid (pointing error exaggerated for illustration); (d) the IMU calibration, with the gravity vector $\hat{\vec{g}}$ (yellow) and the ground plane normal (magenta) before (top) and after (bottom) calibration, which minimizes the angle between them.}
\label{Fig:input}
\end{figure*}

\subsection{Range Sensing}\label{Sec:range_sensing}
We focus on \textit{organized} point cloud data~\cite{RC11} in
the form of an image grid, acquired from a single point of view,
either taken directly from a depth camera ({$640~\times~480$}
clouds in $30~{\rm Hz}$, either from a Microsoft Kinect or a
Primesense Carmine 1.09, depending on the application's range
limits, as appears in Fig.~\ref{Fig:input}-(a,b)) or indirectly
(Sec.~\ref{Sec:track_amvkf}) from a volumetric map, where depth
images are fused over space and time.  The 3D point coordinates
$(x,y,z)$ in the camera frame, where the $\vec{x}$-axis
points to the right and the $\vec{y}$-axis down in the camera
image, can be expressed as a function of the coordinates of the
measurement ray direction vector $\vec{m} = (m_x,m_y,m_z)$ through pixel
$(u,v)$ and the range $r$ of the point along that
vector as (Fig.~\ref{Fig:input}-(c)):
\begin{equation}
 [x\ y\ z] = [m_x\ m_y\ m_z]\ r \label{Eq:mr2xyz}
\end{equation}

The acquired point cloud data uncertainty can be either due to
sensor inaccuracies or due to triangulation errors (the
correspondence problem~\cite{SS02}) and can be split into two
categories: outliers and noisy inliers.  Detecting and removing
outliers is extensively studied and thus a set of real-time
pre-processing filters, such as discontinuity-preserving
bilateral filter, are applied first.  Furthermore, as we will see
in Sec.~\ref{Sec:tracking}, KinectFusion inherently ignores some
outliers, when fusing data over time.

There are various ways to quantify the uncertainty of noisy
inlier points, i.e.~points that deviate from the underlying
ground truth surface.  In~\cite{VK11}, we express the 3D point
uncertainty using Gaussian modeling, with {$3~\times~3$}
covariance matrices.  To estimate these matrices, we experimented
with constant, linear, and quadratic error model assumptions, but
we preferred Murray and Little's~\cite{ML05} two-parameter error
model for stereo disparity uncertainty, which experimentally
appears to follow better the distribution of the input point
cloud data.  The error model is represented by two non-negative
parameters: the variance of the pointing error of the measurement
vectors $\sigma_p$ and the disparity matching error $\sigma_{\vec{m}}$,
both measured in pixels. The covariance matrix for a 3D point in
physical units is
	\begin{eqnarray}
		\Sigma = J E J^T, \
		E =  \begin{bmatrix}
	  		 \sigma_p & 0 & 0 \\
  			 0 & \sigma_p & 0 \\
  			 0 & 0 & \sigma_{\vec{m}}
 		     \end{bmatrix}, \
 		J =  \begin{bmatrix}
	  		 \frac{b}{d} & 0 & -\frac{b u}{d^2} \\
  			 0 & \frac{b}{d} & -\frac{b v}{d^2} \\
  			 0 & 0 & -\frac{f_x b}{d^2}
 		     \end{bmatrix}
	\end{eqnarray}
where $b$ is the stereo baseline (in physical units),
$d$ is the disparity (in pixels), $(u,v)$ are the
image pixel coordinates, and $f_x$ the depth camera focal
length (in pixels).  The error model parameters we used for the
Kinect are: $\sigma_p = 0.35$px, $\sigma_{\vec{m}} = 0.17$px; the former is
from~\cite{KM10}, the latter we determined experimentally
following~\cite{ML05}.  As can been also seen in
Fig.~\ref{Fig:input}-(c), range data exhibit heteroskedasticity
(non-uniform variance)---typically, there is much more
uncertainty in range than aim~\cite{PVB09,ML05} and the variance
changes with range. Since the measurement rays usually have a
single center of projection, the error ellipsoids for the sampled
points are not co-oriented, because each is elongated in the
direction of its own measurement ray.

\begin{figure*}
\includegraphics[width=\textwidth]{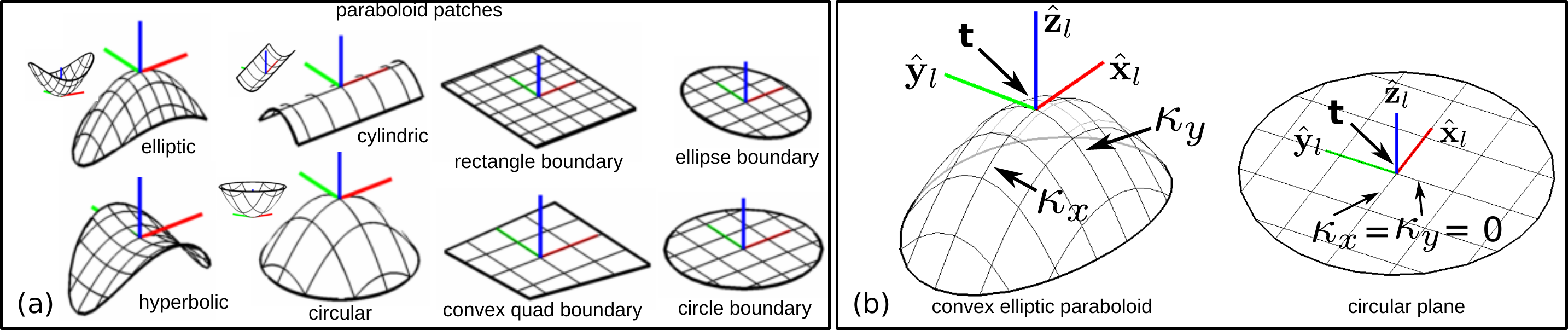}
\caption{{(a)} Examples of eight (convex) paraboloid patch types, including the local coordinate frames and the concave variants shown inset. {(b)} The intrinsic and extrinsic parameters for the elliptic paraboloid and the circular plane patches.\label{Fig:patches}}
\end{figure*}

\subsection{Inertial Measurement Unit ({IMU})}
The use of proprioceptive Inertial Measurement Units (IMUs) for sensing the direction of gravity, is very useful for locomotion.  Using a CH Robotics UM6 $9$-DoF IMU, mounted on the top of our range sensors (Fig.~\ref{Fig:input}-(a)), we receive $100~{\rm Hz}$ IMU data, spatiotemporally coregistered with the $30~{\rm Hz}$ RGB-D data, received from the depth sensor.  Temporal registration of the RGB, depth, and IMU datastreams is based on timestamps.  Spatial registration of the RGB and depth data is based on manufacturer hardware calibration and image warping, implemented in the hardware driver.

Our depth camera and IMU are rigidly attached together.  Precise
spatial registration of the depth and IMU data uses a calibration
algorithm, that we implemented to calculate the rotation of the
UM6 relative to the range sensor, from a dataset of depth images
of a flat horizontal surface (a flat floor) and the corresponding
UM6 orientation data.  The gravity vector $\hat{\vec{g}}$ for the UM6
is calculated directly from its orientation data.  We pair each
gravity vector with the corresponding one in the depth camera
coordinate frame, estimated as the downward facing normal of the
dominant plane.  For all these pairs of gravity vectors we solve
the orthogonal Procrustes problem~\cite{ELF97}, that gives the
best-fit IMU-to-depth-camera rotation (Fig.~\ref{Fig:input}-(d)).

 \section{Surface Modeling using Curved Patches}
 \label{Sec:patches}
Modeling the shape and pose of contacts for locomotion on an
unstructured trail, is still an open problem in robot vision.
Inspired by natural systems like humans, we introduce a set of
patch models, to sparsely model potential contacts.  We first
review the patch modeling, fitting, and validation process that
we previously introduced in~\cite{VK11,KV13,KV14}.
%\QUERY[7]

\subsection{Patch Modeling} \label{Sec:patch_modeling}

\begin{table}[h]
\begin{center}
{\setlength{\tabcolsep}{2pt}
\begin{tabular}{|l|l|r|r|c|l|}\hline
{\bf surface}         & {\bf bound} & \multicolumn{2}{|l|}{\bf parameters}   & {\bf DoF}             & {\bf curvature/boundary}\\\cline{3-4}
                      &             & {\bf intrin.}                          & {\bf extrin.}         &           & {\bf parameters}\\\hline\hline
elliptic parab.       &ellipse&$\vec{d}_e,\vec{k}$ &$\vec{r},\vec{c}$     &10&\,$\sgn(\kappa_x)\!\!=\!\!\sgn(\kappa_y)$\\\hline
hyperbolic parab.     &ellipse&$\vec{d}_e,\vec{k}$ &$\vec{r},\vec{c}$     &10&\,$\sgn(\kappa_x)\!\!\neq\!\!\sgn(\kappa_y)$\\\hline
cylindric parab.      &aa rect&$\vec{d}_r,\kappa$  &$\vec{r},\vec{c}$     &9 &\,$\vec{k}\!\!=\!\![0\ \kappa]^T$\\\hline
circular parab.       &circle &$d_c,\kappa$    &$\vec{r}_{xy},\vec{c}$&7 &$\vec{k}\!\!=\!\![\kappa\, \kappa]^T$\!\!,$\vec{d}_e\!\!=\!\![d_c\, d_c]^T$\\\hline
\multirow{4}{*}{plane}&ellipse&$\vec{d}_e$         &$\vec{r},\vec{c}$     &8 &\,$\vec{k}\!\!=\!\!\vec{0}$\\\cline{2-6}
                      &circle &$d_c$               &$\vec{r}_{xy},\vec{c}$&6 &\,$\vec{k}\!\!=\!\!\vec{0}$, $\vec{d}_e\!\!=\!\![d_c\, d_c]^T$\\\cline{2-6}
                      &aa rect&$\vec{d}_r$         &$\vec{r},\vec{c}$     &8 &\,$\vec{k}\!\!=\!\!\vec{0}$\\\cline{2-6}
                      &c quad &$\vec{d}_q$         &$\vec{r},\vec{c}$     &11&\,$\vec{k}\!\!=\!\!\vec{0}$\\\hline
sphere                &circle &$d_c,\kappa$        &$\vec{r}_{xy},\vec{c}$&7 &\,$\vec{d}_e\!\!=\!\![d_c\, d_c]^T$\!\!, $|\kappa|d_c\!\!\leq\!\!1$\\\hline
circular cylinder     &aa rect&$\vec{d}_r,\kappa$  &$\vec{r},\vec{c}$     &9 &\,$|\kappa|d_y\!\!\leq\!\!1$\\\hline
\end{tabular}
}
\end{center}
\caption{The 10 bounded curved patch types.}%{The 10 patch types.}
\label{tb-patches}
\end{table}

We introduce a set of ten curved/flat patch types (Table~\ref{tb-patches}; eight of them are shown in Fig.~\ref{Fig:patches}-(a)), suitable to model rough surfaces and to balance expressiveness with compactness of representation: eight are first and second-degree polynomials---paraboloids and planes.  Two are non-paraboloids -- spherical and cylindrical -- to better approximate smooth surfaces, particularly on man-made structures (e.g.~pipes and railings).  We pair each surface type with a specific boundary curve to capture useful symmetries and asymmetries.  This particular system is not the only possible taxonomy, but it reflects our design choices in an attempt to balance expressiveness vs minimality.

\subsubsection{Extrinsic and Intrinsic Surface Parameters}
 \label{Sec:params}
 
\begin{figure*}
\includegraphics[width=\textwidth]{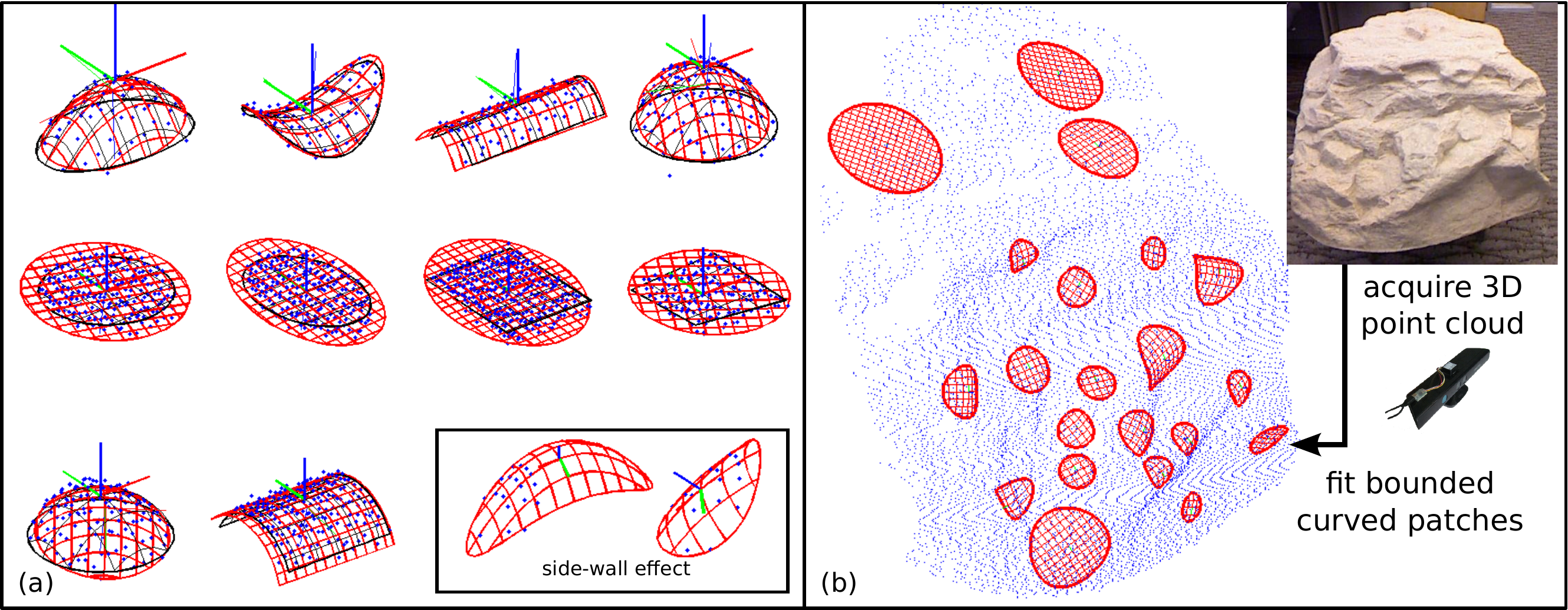}
\caption{{(a)} Automatic fits in red for all patch types,
with simulated noisy range samples in blue; the ``side-wall''
effect reparameterization, shown inset (see~\cite{KV13}).
(b) A set of 21 patches, manually chosen and
automatically fit to the samples taken from a rock model, using
an RGB-D Kinect sensor. \label{Fig:patch-fitting}}
\end{figure*}

We parametrize a patch with a vector of real parameters that
describe the shape (curvature and boundary \emph{intrinsic}
parameters) and 3D rigid-body pose (\emph{extrinsic} parameters).
The parametrization should be done in a way that it is minimal
and also enables separate uncertainty modeling for the intrinsic
and extrinsic parameters, considering the \emph{continuous
symmetry class} of the patch~\cite{Srinivasan03}.  Grassia showed
in~\cite{Gra98} that a pose representation that is beneficial for
iterative optimization methods (as our patch fitting described in
Sec.~\ref{Sec:patch_fitting}) is:
\beg
  [\vec{r}^T\ \vec{t}^T]^T\in\R^6\text{ with }(\vec{r},\vec{t})\in\R^3\times\R^3 \label{Eq:rt}
\eeg
where $\vec{t}$ is a translation and $\vec{r}$ is an \emph{orientation vector}, giving an element of $SO(3)$ via an exponential map.  We use Rodrigues' rotation formula for the exponential map $R(\vec{r})\!:\!\R^3\!\rightarrow\!SO(3)\!\subset\!\R^{3\times3}$ (Grassia used quaternions):
\begin{gather}
  R(\vec{r})=I+[\vec{r}]_\times\alpha+[\vec{r}]_\times^2\beta \label{Eq:rexp}\\
\label{Eq:R}
  \theta\defeq\| \vec{r}\| ,\ 
  \alpha\defeq\frac{\sin\theta}{\theta},\ 
  \beta\defeq\frac{1-\cos\theta}{\theta^2}\nonumber\\
  \vec{r}=\left[
\begin{smallmatrix}
r_x\\r_y\\r_z
\end{smallmatrix}
\right],\ 
  [\vec{r}]_\times\defeq
  \left[
\begin{smallmatrix}
0&-r_z&r_y\\r_z&0&-r_x\\-r_y&r_x&0
\end{smallmatrix}
\right].\nonumber
\end{gather}
Algebraically, $(\vec{r},\vec{t})$ corresponds to a {$4~\times~4$} homogeneous rigid body transformation of $SE(3)$:
\[
\begin{bmatrix}
R(\vec{r})&\vec{t}\\\vecmb{0}^T&1
\end{bmatrix}
\]
The pose of a patch can be thus defined as the pose of a local coordinate frame $L$ relative to a world frame $W$, where $R(\vec{r})$ is a basis for $L$ and $\vec{t}$ is its origin.  We define the familiar functions $X_{f,r}:\R^3\times\R^3\times\R^3\rightarrow\R^3$ that transform a point $\vec{q}_l$ in $L$ to/from the corresponding $\vec{q}_w$ in $W$:
\begin{align}
  \vec{q}_w &= X_f(\vec{q}_l,\vec{r},\vec{t}) \defeq R(\vec{r})\vec{q}_l+\vec{t} \label{Eq:xf}\\
  \vec{q}_l &= X_r(\vec{q}_w,\vec{r},\vec{t}) \defeq R(-\vec{r})(\vec{q}_w\!\!-\!\vec{t})=R(\vec{r})^T(\vec{q}_w\!\!-\!\vec{t}) \label{Eq:xr}
\end{align}
where (\ref{Eq:xr}) makes use of the inverse transform
\begin{equation}\label{eq-xinv}
  (\vec{r},\vec{t})^{-1}\defeq(-\vec{r},-R(-\vec{r})\vec{t})=(-\vec{r},-R(\vec{r})^T\vec{t}).
\end{equation}
The 6 DoF patch pose parametrization is defined from Eqs.~(\ref{Eq:rt})--(\ref{Eq:xr}).  For the 5 DoF pose representation in the case of rotationally symmetric patches, we fix $r_z$ at 0 in (\ref{Eq:rt})--(\ref{Eq:xr}):
\begin{equation}\label{Eq:rt2}
  (\vec{r}_{xy},\vec{t})\in\R^2\times\R^3\ \text{corresp.}\ ([\vec{r}_{xy}^T\ 0]^T,\vec{t})\in\R^3\times\R^3.
\end{equation}
We now, present the details of the elliptic and hyperbolic paraboloid patch models, which are two out of the ten proposed surfaces. Due to space constraints we must refer readers to~\cite{VK11} for further details on: (1) the intrinsic and
extrinsic parametrization in the case of symmetries, (2) the
methodology to avoid singularities, by a fast dynamic
reparameterization, and (3) the circular cylindrical and
spherical patch models.

\subsubsection{Paraboloids}
 \label{Sec:paraboloids}
The second degree polynomial, that locally fits a smooth surface $S\subset\R^3$ at a given point $\vec{t}\in S$, is called a paraboloid.  It is an one-sheet quadric, with a central point of symmetry ($\vec{t}\in S$) and two independent curvatures $\kappa_x, \kappa_y$ (principal curvatures of $S$ at $\vec{t}$) in orthogonal directions (Fig.~\ref{Fig:patches}-(b)).

We define $\uvec{x}_l$ and $\uvec{y}_l$ to be the unit vectors in
the directions of the principal curvatures in the tangent plane
to $S$ at $\vec{t}$ and $\uvec{z}_l\defeq\uvec{x}_l\times\uvec{y}_l$ the surface normal
to $S$ at $\vec{t}$.  In a world coordinate frame
$W$, $\vec{t}\in\R^3$ is the origin of $S$ and
$R\defeq[\uvec{x}_l\ \uvec{y}_l\ \uvec{z}_l]$ is its local frame $L$ at
$\vec{t}$~\cite{HS02}.  A general paraboloid can be
parametrized by $\vec{k}\defeq[\kappa_x\ \kappa_y]^T$,  $\vec{r}$, and $\vec{t}$,
while with the use of the log map, i.e.~$(\vec{r},\vec{t})\defeq(\vec{r}(R),\vec{t})$, points can
be transformed from $L$ to $W$.

The implicit and explicit forms for the paraboloid in standard position in $L$ are:
\begin{align}
  0=p_{li}(\vec{q}_l,\vec{k})&\defeq\vec{q}_l^T\diag([\vec{k}^T\ 0]^T)\vec{q}_l-2\vec{q}_l^T\uvec{z} \label{Eq:pli}\\
  \vec{q}_l=p_{le}(\vec{u},\vec{k})&\defeq[\uvec{x}\ \uvec{y}]\vec{u}+\frac{1}{2}\vec{u}^T \diag(\vec{k})\vec{u}\uvec{z} \label{Eq:ple}
\end{align}
\begin{gather}
p_{li}\!:\!\R^3\!\times\!\R^2\!\!\rightarrow\R,\
p_{le}\!:\!\R^2\!\times\!\R^2\!\!\rightarrow\R^3
\nonumber
\end{gather}
where $\vec{q}_l\in\R^3$ is a point on the patch in $L$ and $\vec{u}\in\R^2$ parameters of the explicit form.  Composing (\ref{Eq:pli}), (\ref{Eq:ple}) with (\ref{Eq:xf}), (\ref{Eq:xr}), one can describe $\vec{q}_w\in\R^3$ in world frame $W$:
\begin{align}
  0=p_{wi}(\vec{q}_w,\vec{k},\vec{r},\vec{t})&\defeq p_{li}(X_r(\vec{q}_w,\vec{r},\vec{t}),\vec{k}) \label{Eq:pwi}\\
  \vec{q}_w=p_{we}(\vec{u},\vec{k},\vec{r},\vec{t})&\defeq X_f(p_{le}(\vec{u},\vec{k}),\vec{r},\vec{t}) \label{Eq:pwe}
\end{align}
\begin{gather}
  p_{wi}:\R^3\!\!\times\!\!\R^2\!\!\times\!\!\R^3\!\!\times\!\!\R^3\rightarrow\R,\ p_{we}:\R^2\!\!\times\!\!\R^2\!\!\times\!\!\R^3\!\!\times\!\!\R^3\rightarrow\R^3 \nonumber
\end{gather}
where $\vec{u}$ is the projection of $\vec{q}_l$ onto the local frame $xy$ plane:
\begin{equation}\label{Eq:u}
  \vec{u}\defeq\Pi_{xy}\vec{q}_l=\Pi_{xy}X_r(\vec{q}_w,\vec{r},\vec{t}),\ \Pi_{xy}\defeq
  [\uvec{x}\ \uvec{y}]^T.
\end{equation}

\subsubsection*{Boundaries}

We bound elliptic and hyperbolic paraboloid patches with ellipses in the $xy$ plane of the local frame $L$, axis aligned and centered at $\vec{t}$ (Fig.~\ref{Fig:patches}).  Defining the ellipse radii $\vec{d}_e\defeq[d_x\ d_y]^T$, the bounded patch is the subset of the full surface (\ref{Eq:pli})--(\ref{Eq:pwe}) where $e\!:\!\R^2\!\times\!\R^2\!\!\rightarrow\R$ and $\vec{u}$ satisfies:
\begin{equation}\label{Eq:ellipse}
  0\geq e(\vec{u},\vec{d}_e)\defeq\vec{u}^T\diag([1/d_x^2\ 1/d_y^2])\vec{u}-1.
\end{equation}

Our boundary definitions for the cylindric and circular paraboloids, the rectangle, ellipse, circle, and convex quad bounded planes, as well as the spherical and circular cylindrical bounded patch models can be found in~\cite{VK11}.

\subsection{Patch Fitting} \label{Sec:patch_fitting}
\begin{figure*}
\includegraphics[width=\textwidth]{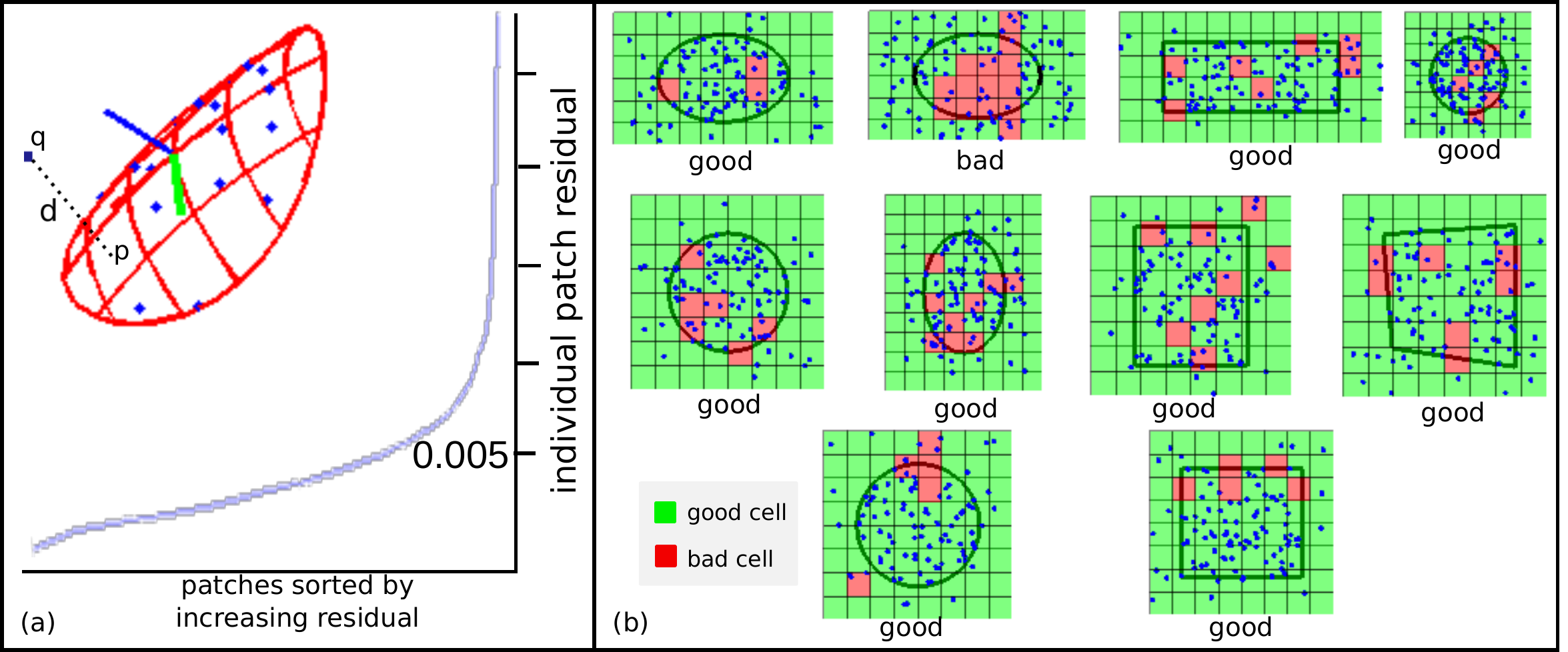}
\caption{{(a)} Residual evaluation, with an example of the Euclidean distance $d$ between a point $\vec{q}$ and its projection on the patch and a plot with 1000 residuals of patches fitted randomly on the rock model in (Fig.~\ref{Fig:patch-fitting}-(b)). {(b)} Coverage evaluation, for all patch types in simulated data. \label{Fig:patch-fit-eval}}
\end{figure*}

Having defined the patch models, we now present our method to fit patches (Fig.~\ref{Fig:patch-fitting}), by solving the non-linear $\chi^2$ maximum likelihood problem when: (1) data points are corrupted by heteroskedastic (nonuniform variance) Gaussian noise and (2) bounded paraboloid patches need to be fit, not just unconstrained general quadrics.

In~\cite{VK11} we give a non-linear fitting algorithm, which is
based on a variation of Levenberg--Marquardt iteration, called
Weighted Levenberg--Marquardt (WLM).  WLM minimizes a
sum-of-squares residual, by optimizing the patch implicit and
explicit parameters.  The residual for an individual sample point
is computed by scaling the value of the implicit form by the
inverse of a first-order estimate of its standard deviation,
which is derived in turn from a covariance matrix, modeling the
sensor uncertainty for the point.  Due to space constraints we
must refer readers to the original paper for the full description
of the WLM fitting process.  Here we restrict to the case of
elliptic and hyperbolic paraboloid fitting; the full algorithm
for all patch types is given in~\cite{VK11}.

The {\bf inputs} of the algorithm are: (1) $N$ sample points $\vec{q}_i\in\R^3$, with covariance matrices $\Sigma_i\in\R^{3\times3}$ (positive semi-definite), (2) the general surface type to fit $s\in\; $\{\emph{parab}, plane, sphere, cylinder\}, (3) the boundary type $b\in\; $\{ellipse, circle, rectangle, quadrilateral\}, if $s=\text{plane}$, and (4) a boundary containment probability $\Gamma\in(0,1]$.  The \textbf{outputs} are: (1) the combined vector of patch intrinsic and extrinsic parameters $\vec{p}\in\R^p$, where $p$ is the DoF of the patch type, and (2) the parameter covariance matrix $\Sigma\in\R^{p\times p}$, modeling the uncertainty of the patch.

 For elliptic and hyperbolic paraboloids bounded by ellipses, $p=10$ and $\vec{p}\defeq\left[\vec{d}_e,\vec{k},\vec{r},\vec{t}\right]\in\R^{10}$.  $\vec{d}_e\in\R^2$ gives the ellipse boundary radii; $\vec{k}\in\R^2$ gives the patch curvatures, $\vec{r}\in\R^3$ gives the patch orientation in space as an exponential map vector, and $\vec{t}\in\R^3$ gives the location of the patch apex in space.  $\vec{d}_e,\vec{k}$ are the intrinsic and $\vec{r},\vec{t}$ the extrinsic parameters of the patch.

The algorithm proceeds in 2 stages, which include heuristics for
avoiding local minima, when solving the non-linear system.  The
first three steps fit an unbounded surface; the rest are largely
concerned with fitting the boundaries, which can include final
resolution of the patch center and orientation, where the
bounding shape breaks symmetries of the underlying surface.
Principal Component Analysis (PCA) is applied to the 2D projected
points onto the local $xy$-plane, for finding the
enclosing boundary.  A closed-form solution~\cite{RVC02} for the
corresponding eigendecomposition problem using 1D and 2D moments
to fit approximate boundaries is used.

\subsubsection{Paraboloid Patch Fitting Algorithm} \label{Sec:fit}

\noindent {\bf Fit Patch Surface}
  
\noindent \textit{Step 1: Plane fitting} 

 Fit a plane with linear least-squares, ignoring $\Sigma_i$ and set $\vec{t}\gets\bar{\vec{q}}-\uvec{z}_l^T(\bar{\vec{q}}-\vec{t})\uvec{z}_l$ (perpendicular projection of $\bar{\vec{q}}\defeq\avg(\vec{q}_i)$ on plane).
    
\noindent \textit{Step 2: Surface Fitting} 

 With $\vec{k}=[0\ 0]^T$, $\vec{r}\defeq[\vec{r}_{xy}^T\ 0]^T$, and $\vec{t}$ from step~1 as initial estimates fit an unbounded paraboloid using WLM on Eq.~(\ref{Eq:pwi}).

\noindent \textit{Step 3: Curvature Discrimination} 

 Refine the patch based on the fitted curvatures
$\vec{k}=[\kappa_x\ \kappa_y]^T$.  The patch is an elliptic paraboloid if and only if
$\sgn(\kappa_x)=\sgn(\kappa_y)$, otherwise change the type of the patch and the
fitting process as described in~\cite{VK11}.

\noindent {\bf Fit Patch Boundary}

\noindent {Step 4: Initialize Bounding Parameters}
    
  Set $\lambda \defeq \sqrt{2}\erf^{-1}(\Gamma)$ for boundary containment
scaling~\cite{Ribeiro04}. Project the data $\vec{q}_i\in\R^3$ to
$\vec{u}_i=[x_i\ y_i]^T\in\R^2$ using (\ref{Eq:u}).
    
    Set first and second data moments: $\bar{x}\defeq\avg(x_i)$, $\bar{y}\defeq\avg(y_i)$, $v_x\defeq\avg(x_i^2)$, $v_y\defeq\avg(y_i^2)$.

\textit{Step 5: Ellipse Boundary Fitting}
    
Set $\vec{d}_e=\lambda[\sqrt{v_x}\ \sqrt{v_y}]^T$.

The full algorithm in~\cite{VK11}, has $9$ steps to deal
with all our curved bounded patch types.

\subsection{Patch Validation} \label{Sec:validation}
Evaluating if a patch faithfully represents the surface after the
fitting process, is a necessary task.  For example, the iterative
WLM part of the fitting algorithm may diverge or get stuck in a
local minimum, or the underlying data may be cluttered and not
cover a major part of the patch.  We review three measures that
we introduced in~\cite{KV13}, based on: (1) \emph{residual} and
\emph{curvature}, to evaluate the patch surface shape and (2)
\emph{coverage}, to evaluate its boundary
(Figs.~\ref{Fig:patch-fit-eval},~\ref{Fig:patch-tot-eval}).
Further experimental results in the fitting and validation are
presented in Sec.~\ref{Sec:exp}.

\subsubsection{Residual Evaluation}
 \label{Sec:patch_residual}
The patch residual (Fig.~\ref{Fig:patch-fit-eval}-(a)), measures the
deviation between the sample points and the unbounded patch
surface.  The deviation is a result of noise in the data and/or
local minima during the WLM fitting process.  The residual is the
Euclidean root-mean-square error (\emph{RMSE}~\cite{PKPSTV93})
$\varrho$, between the sample points $\vec{q}_i$ and their
corresponding closest points $\vec{p}_i$ on the patch:
\begin{equation}
 \varrho = RMSE(\{\vec{q}\},\{\vec{p}\}) = 
  \sqrt{\frac{\sum_{i=1}^N{\| \vec{q}_i- \vec{p}_i\| ^2}}{N}},\label{Eq:rmse}
\end{equation}
where $N$ is the total number of points.  Calculating
the $\vec{p}_i$ for each $\vec{q}_i$, is a computationally
expensive problem and is solved either using Lagrange
multipliers~\cite{Eberly1999} or an approximation such as
Taubin's~\cite{Taubin91}.

The residual threshold $T_r$, such that patches with
$\varrho > T_r$ are dropped, was determined experimentally, by
fitting $1000$ random patches of $r=0.1$ m radius on a
rock dataset~\cite{KV13} and sorting them with respect to their
residuals.  The value $T_r=0.01$ m was selected to include
approximately $95\%$ of the patches.  This is not the only
way to define $T_r$, which could be a task dependent
threshold.  For instance, $\varrho_{alt} = \max\| \vec{q}_i-\vec{p}_i\| $ could be used to check if
any small surface bumps protrude more than a desired amount.
Fig.~\ref{Fig:patch-tot-eval} (patches in purple), illustrates
patches that pass and fail the residual evaluation.

\subsubsection{Coverage Evaluation}
 \label{Sec:patch_coverage}
An evaluation that takes into account the patch boundary is also needed to detect the cases that too many sample points are outside the patch boundary or too many areas in the boundary are not supported by data.  In these cases, the patch may not faithfully represent the data.  We handle this, by generating an axis-aligned cell grid of fixed pitch $w_c$ on the $xy$ plane of the patch local frame $L$ (Fig.~\ref{Fig:patch-fit-eval}-(b) illustrates 2D-projected patches and cells that pass and fail the coverage evaluation).

Let $I_c$ (resp. $O_c$) be the number of data points, whose $xy$ projections is in cell $c$ and inside (resp. outside) the projected patch boundary and $A_i$ the area of the geometric intersection of the cell and the projected patch boundary.  We consider a cell to be bad  if and only if
\begin{equation}
\label{Eq:cov}
 I_c < \frac{A_i}{w_c^2} T_i \text{ or } O_c > (1-\frac{A_i}{w_c^2})T_o,
\end{equation}
for thresholds $T_i$ and $T_o$.  We fix these thresholds relative to the expected number of samples $N_e$ in the ideal scenario, where sample points are evenly distributed inside the patch boundary:
\begin{equation}
  T_i = \zeta_i N_e,\ \ T_o = \zeta_o N_e,\ \ N_e\defeq N/N_p,\ \ N_p\defeq\frac{A_p}{w_c^2}, 
\end{equation}
where $N$ is the total number of sample points and $A_p$ is the area of the patch, approximated as the area inside the projected boundary.  A patch is bad if and only if there are more than $T_p$ bad cells.  We determine experimentally the threshold values, after fitting several patches of $r=0.1$ m radius to $w_c = 0.01$ m, $\zeta_i = 0.8$, $\zeta_o = 0.2$, and $T_p = 0.3 N_p$.  Fig.~\ref{Fig:patch-tot-eval} (patches in green), illustrates patches that pass and fail the coverage evaluation.

\subsubsection{Curvature Evaluation}
\label{Sec:curvature_validation}
\begin{figure}
\includegraphics{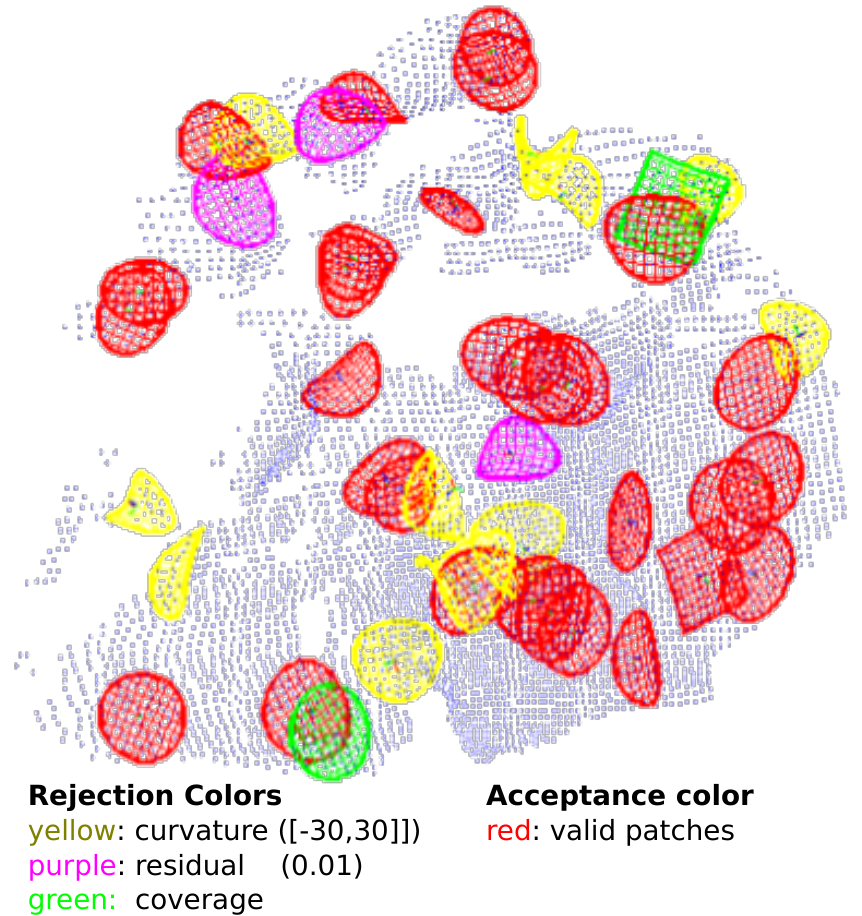}
\caption{Red patches that passed the validations and yellow/purple/green patches that failed curvature/residual/coverage evaluation, respectively. \label{Fig:patch-tot-eval}}
\end{figure}

There may be cases, where residual and coverage may pass, but the
bounded patch does not represent the data correctly.  This may
happen either when the point samples are acquired from very
curved surfaces or when the LM non-linear fitting gets stuck in
local minima.  For this reason a patch will be marked bad
if and only if its minimum curvature is smaller
than a threshold $\kappa_{min,t}$ or its maximum curvature is bigger
than a threshold $\kappa_{max,t}$.   Like previously, we set these
thresholds experimentally; for details see~\cite{KV13}.\junk{to
$\kappa_{min,t} = -1.5\max(\vec{d})$ and $\kappa_{max,t} = 1.5\max(\vec{d})$, where $\vec{d}$ is the patch
boundary vector.}  Fig.~\ref{Fig:patch-tot-eval} (patches in yellow),
illustrates patches that pass and fail the curvature evaluation.

\section{Spatial Patch Mapping}
 \label{Sec:mapping}

 In the previous sections, we described the framework for: (1) modeling contact patches between a robot and local surfaces in the environment and (2) fitting them to 3D point cloud data.  We now, describe a five-stage algorithm to find and spatially map potentially useful patches.  Each stage is described as a batch operation, but the last three stages may also work in such a way that patches are added to the map incrementally until a time or space limit is met.  The patch mapping is dovetailed with the patch tracking, introduced in the next section, for maintaining a spatially and temporally coherent map of  nearby patches, as the robot moves through the environment:
\begin{enumerate}[align=left]
  \item[{{\bf Stage i:}}] Acquire RGB-D and IMU Data (Sec.~\ref{Sec:map_input}).
  \item[{{\bf Stage ii:}}] Preprocess the Input Point Cloud (Sec.~\ref{Sec:map_preprocess}).
 \item[{{\bf Stage iii:}}] Select Seed Points on the Surface (Sec.~\ref{Sec:map_seeds}).
 \item[{{\bf Stage iv:}}] Find Seed Point Neighborhoods (Sec.~\ref{Sec:map_neighborhood}).
 \item[{{\bf Stage v:}}] Fit Patches to the Neighborhoods (Sec.~\ref{Sec:map_patch_model_fit}).
\end{enumerate}

A homogeneous spatial patch map (Sec.~\ref{Sec:spatial_map}) is built, as follows.  First, RGB-D and IMU data are acquired, as explained in Sec.~\ref{Sec:input} (\textbf{Stage i}), followed by data preprocessing, such as saliency filtering or decimation, depending on the application (\textbf{Stage ii}).  Then, a set of seed points (\textbf{Stage iii}) and their $r$-size neighborhoods (\textbf{Stage iv}) are found for fitting and validating patches (\textbf{Stage v}).  We fix $r$ to match the size of the intended contact surface on the robot, such as the sole of the foot.

We first introduce the \emph{local volumetric workspace} (or simply the \emph{volume}), which will be the moving local working space around the robot.

 \subsection*{Local Volumetric Workspace}
 \label{Sec:Volume}
For a moving robot in the environment, 3D point cloud and IMU data are acquired at relatively high frame rates (typically 30--100~Hz).  Information fusion is desirable not only to remove redundancies over time, but also to have a terrain representation of areas that are sometimes occluded or obstructed by the robot itself.  Only the area around the robot is required for local 3D fine contact planning---longer-range perception could be used separately for coarse trajectory planning and obstacle avoidance.

Thus, it is natural here to consider only the data in a moving
volume around the robot.  We define such a volume, as a cube with
a \emph{volume coordinate frame} at a top corner
(Fig.~\ref{Fig:tracking}-(ii)).  We align its $\vec{y}$-axis point
with the gravity vector, derived from the IMU data, and we let
the $\vec{x}$ and $\vec{z}$-axes point along the cube
edges, forming a right-handed frame.  In this way our \emph{local
volumetric workspace} is the same as the Truncated Signed
Distance Function (TSDF) volume in moving volume
KinectFusion~\cite{RV12}.  At any time $t$, the volume
is fully described by: (1) its \textbf{size} $V_s$ (a
predefined constant), and (2) its \textbf{pose} relative to the
camera, with the following {$4~\times~4$} rigid body
transformation from the camera to the volume frame:
\begin{gather}\label{Eq:Vpose}
    C_t = 
\begin{bmatrix}
 R_t & \vec{t}_t\\
                          0   & 1
\end{bmatrix}
  \end{gather}
where $R_t$ is a rotation matrix and $\vec{t}_t$ a translation vector.

The volume pose relative to the environment, roughly follows along with the robot.  Our strategy is to move it periodically, whenever it deviates from a nominal relative pose to the camera more than a distance $c_d$ or an angle $c_a$ threshold (Sec.~\ref{Sec:tracking}).

 \subsection{Input Data Acquisition}
 \label{Sec:map_input}
In the first stage of patch mapping, RGB-D and IMU data are acquired, as described in Sec.~\ref{Sec:input}:

\begin{enumerate}[align=left]
  \item[{{\bf Stage i:}}] Acquire RGB-D and IMU Data
\begin{enumerate}
      \item[{{\it Step 1:}}] Receive image $Z$ from the depth camera and absolute orientation quaternion $\quat{q}$ from the IMU.  The depth camera may either be a physical sensor, like Kinect, returning {$640~\times~480$} images, or a virtual camera (implemented by raycasting in the TSDF) in the context of KinectFusion (Sec.~\ref{Sec:tracking}), which typically has a lower resolution, e.g.~{$200~\times~200$}.  In the later case, the virtual camera may also have a different pose in the volume than the physical amera (e.g.~synthetic birds-eye view).
      \item[{{\it Step 2:}}] Convert $Z$ to an $M \times N$ organized point cloud $C$ in camera frame and $\quat{q}$ to a unit gravity vector $\uvec{g}$ pointing down in camera frame.
\end{enumerate}
\end{enumerate}

\subsection{Point Cloud Preprocessing} \label{Sec:map_preprocess}
\begin{figure*}
\includegraphics[width=\textwidth]{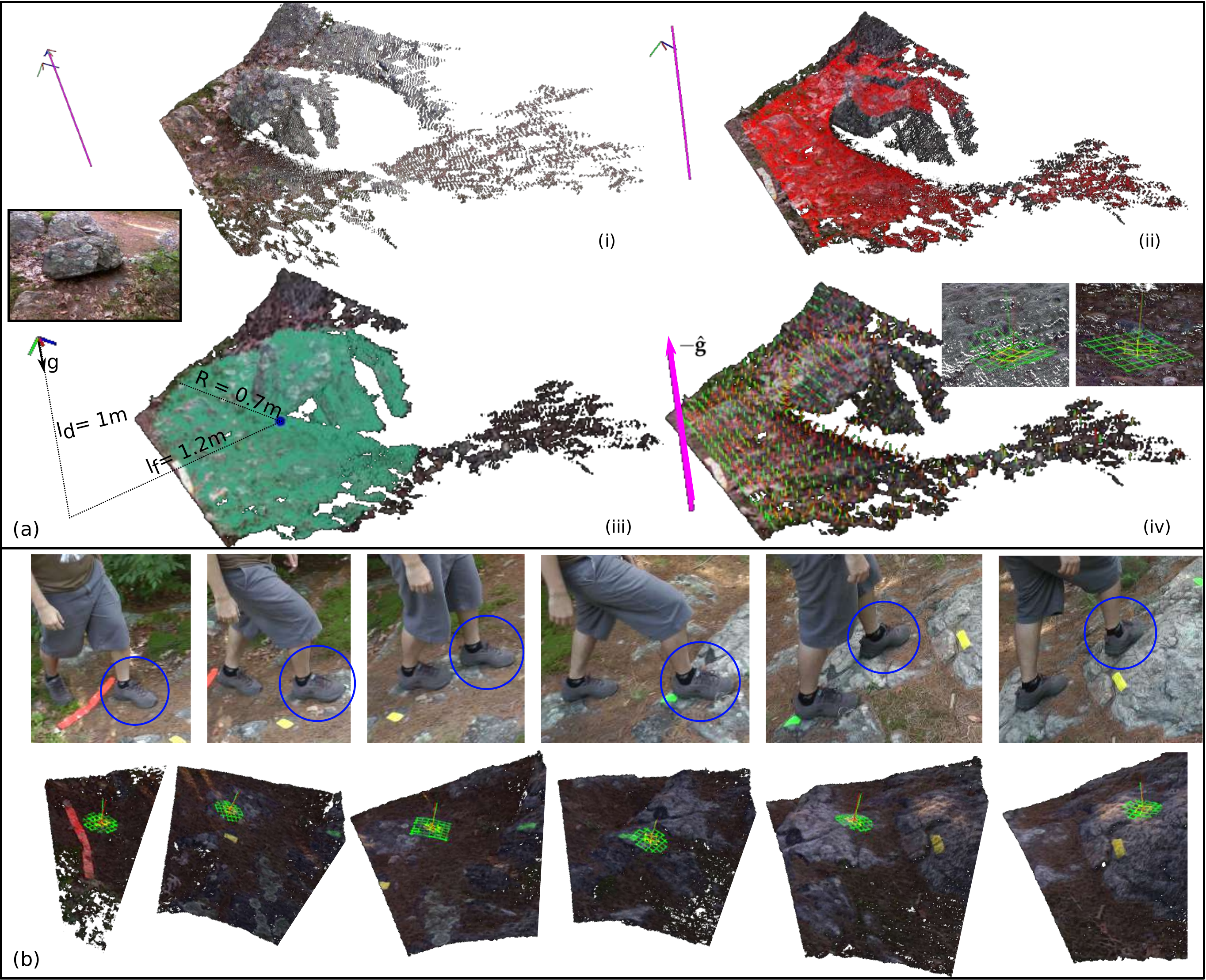}
\caption{{(a)} (i) Dense {$640~\times~480$} RGB-D and IMU data, for an outdoor rock; (ii) salient points in red, after applying DoN and DoNG filtering; (iii) illustration of the DtFP measure, for a $4$ m away fixation point; (iv) illustration of the DoNG and DoN measures, in the two embedded images.  (b) Human selected patches, in RGB-D recordings. \label{Fig:mapping}}
\end{figure*}

Point cloud preprocessing usually depends on the task application requirements.  In this paper, we focus on bipedal locomotion on rough terrain.  Notice, that we restrict our filtering to have real-time performance (i.e.~$\sim$30~Hz) and when points are required to be filtered out, we replace them with NaN values to maintain their organization (used later in optimized methods, e.g.~for fast neighborhood searching).
\begin{enumerate}[align=left]
\item[{\bf Stage ii:}] Preprocess the Input Point Cloud
\begin{enumerate}
      \item[{{\it Step 3:}}] Remove ``background'' points, either by using a filter to threshold the $z$-coordinate values in the camera frame or by setting the volume size $V_s$ appropriately and keeping only the points in it.

   \item[{{\it Step 4:}}] Apply a discontinuity-preserving bilateral filter to
$C$ to reduce outliers~\cite{PF06}.
      
      \item[{{\it Step 5:}}] Optionally, in the case that $C$ comes from a {$640~\times~480$} physical depth camera, downsample $C$ with a {$2~\times~2$} median filter, to create an auxiliary point cloud $D$.  If data from step 1 were acquired from a {$200~\times~200$} virtual camera, then $D \defeq C$.

      \item[{{\it Step 6:}}] Create a new point cloud $H$, by applying the hiking salience filter, on point cloud $D$ (Sec.~\ref{Sec:map_saliency}).
\end{enumerate}
\end{enumerate}
Both the $C$ and $H$ point clouds are kept, since they are used in different steps of the algorithm.  $C$ is used for neighborhood searching and $H$ for seed selection.  Now, we describe the hiking saliency filter, that is applied in Step~6.

\subsubsection{Hiking Saliency Filter}
 \label{Sec:map_saliency}
\begin{figure*}
\includegraphics[width=\textwidth]{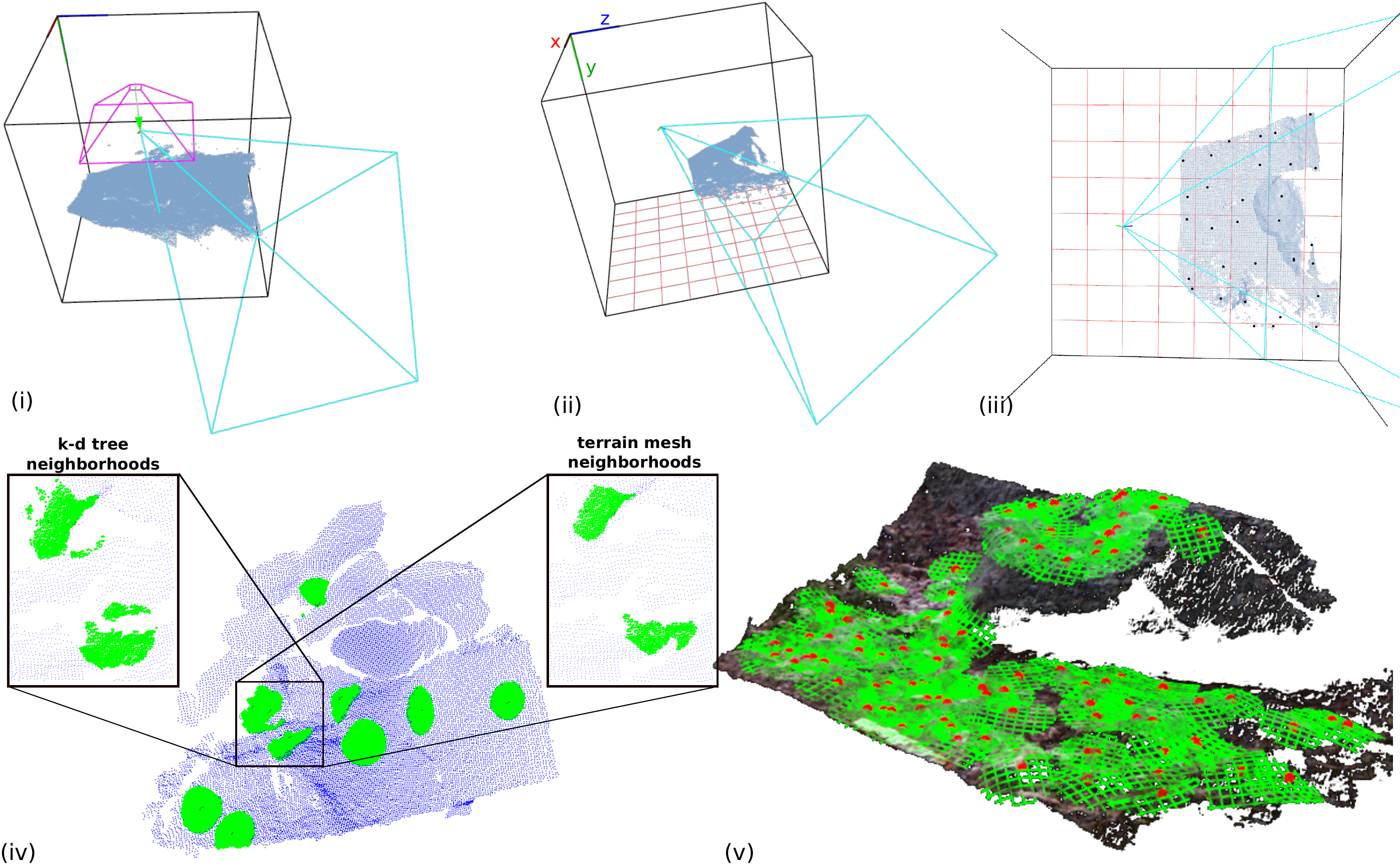}
\caption{The patch mapping system overview.  (i) Point cloud raycasting, from the virtual birds-eye camera (in purple) and the real camera (cyan); (ii) the volume $xy$-plane, divided in an {$8~\times~8$} grid; (iii) one seed per cell, selected randomly, (iv) example of 10 neighborhoods of $r=0.05$ m radius, using the k-d tree and the triangle mesh structures; (v) $100$ patches, randomly fit to salient seeds. \label{Fig:tracking}}
\end{figure*}

In~\cite{KV14}, we describe a real-time bio-inspired system for
automatically finding and fitting salient patches, attempting to
balance patch quality with sufficient sampling of the appropriate
parts of upcoming terrain.  Saliency has been extensively used in
computer graphics (e.g.~\cite{LVJ05}), to describe perceptually
important surfaces.  We introduce three new saliency measures
(Fig.~\ref{Fig:mapping}-(a)), that involve aspects of patch
orientation and location: Difference of Normals (DoN), Difference
of Normal-Gravity (DoNG), and Distance to Fixation Point (DtFP).
They relate to patches that humans commonly select as footholds,
using either bio-mechanical properties, such as that humans tend
to fixate about two steps ahead when they are walking on rough
terrains, or statistical similar patches that we observed humans
to select when hiking (Fig.~\ref{Fig:mapping}-(b)).

\noindent \textbf{Difference of Normals (DoN)}: This operator~\cite{LG12},
relates to the irregularity of a surface at a point, by measuring
the angle between the normals of fine vs coarse scale
neighborhoods.  Points with low DoN can be considered more
salient for footfall selection.  We used the foot size
$r$, for the coarse scale and $r/2$, for the
fine.

\noindent\textbf{Difference of Normal-Gravity (DoNG)}: This operator, relates to the slope of a surface, by measuring the angle between the $r$-neighborhood normal vector of each point and the reverse of the gravity vector $-\hat{\vec{g}}$, acquired by the IMU.  Points with low DoNG can be considered more salient for footfall selection:

\noindent\textbf{Distance to Fixation Point (DtFP)}: This operator,
relates to the biomechanical property of humans to fixate
approximately two steps ahead, when locomoting in rough
terrain~\cite{Marigold08}.  Such a fixation point is estimated
and we consider points that are closer to it, as more
salient.

 These measures are useful to quickly identify good seed points, before fitting, in the following way:
\begin{enumerate}[align=left]
\item[{\bf Stage ii:}] Preprocess the Input Point Cloud
\begin{enumerate}
\item[{Step 6.1}] (\textbf{DtFP}) Using the properties that: $\uvec{g}$ points down and $[1~0~0]^T$ points right in camera frame, estimate the fixation point $\vec{f}$ in camera frame \[\vec{f} \defeq l_d\uvec{g} + l_f([1~0~0]^T \times \uvec{g})\nonumber,
\] where $l_d$, $l_f$ are the distances down and forward, from the camera to the estimated fixation point.
      \item[{Step 6.2}] (\textbf{DtFP}) Initialize $H$, as all the points in $D$ within an $R$-ball region of interest of $\vec{f}$.
   \item[{Step 6.3}] (\textbf{DoN/DoNG}) Compute surface normals
$N,N_s$ corresponding to $D$, using integral
images~\cite{HRDGN12}.  The normal $N(i)$ uses window size
$2rf/Z(i)$, where $Z(i)$ is the $z$ coordinate
(depth)     of point $i$ in camera frame and $f$ is
the focal length of the depth camera in pixels; $N_s(i)$ uses
the half window size.
      \item[{Step 6.4}] (\textbf{DoN}) Remove from $H$ all points $i$, for which \[N(i)^TN_s(i) < \cos(\phi_d),\nonumber
\] where $\phi_d$ is the DoN angle thresholds, estimated from human-selected patches.
      \item[{Step 6.5}] (\textbf{DoNG}) Remove from $H$ all points $i$ for which \[-N(i)^T\uvec{g} < \cos(\phi_g),\nonumber
\] where $\phi_g$ is the DoNG angle thresholds estimated from human-selected patches.
\end{enumerate}
\end{enumerate}
Estimating principle curvatures in real-time, on raw point cloud data, appears to be relatively inefficient.  For this reason we do not apply curvature validation (Sec.~\ref{Sec:curvature_validation}) as a saliency measure in the pre-processing stage.  In contrary, we apply it as a post-processing filtering, on patches that were fit and their curvatures, thus, was determined (Sec.~\ref{Sec:map_patch_model_fit}).

\subsection{Seed Selection} \label{Sec:map_seeds}
Selection of seed points, around which patches will be fit, is an important aspect of our mapping approach.  We uniformly sample the candidate seeds $H$, relative to a coarse grid, imposed on the (horizontal) $xz$-plane in volume frame.  We first split the volume $xz$-plane into $V_{g} \times V_{g}$ grid cells (we typically use $V_g = 8$, see Fig.~\ref{Fig:tracking}-(iii)).  Then, for each cell, we randomly pick up to $n_g$ points, for a total of $n_s$ seed points.  If there are real-time constraints and only a subset of the seeds is needed, we sample first from the cells whose distance from the projected camera position onto the $xz$-plane is smaller (i.e.~closer to the robot), until a time limit is reached.  Moreover, we ignore any new seed points for a cell that already has at least $n_g$ patches, fitted to seeds within it (some of these may be already-existing patches as we describe next).

Notice, that when the volume moves in the physical space, the seeds need to be remapped to new cells.  Since this process may move some prior seeds or patches out of the volume, we remove them also from the map, as described in Sec.~\ref{Sec:tracking}.  If more than $n_g$ patches are remapped into a cell, the extra patches can be culled, if desired.  Fig.~\ref{Fig:tracking}-(iii) illustrates $n_s = 31$ seeds, with the volume divided into an {$8~\times~8$} grid ($V_g = 8$), with one seed point per grid ($n_g = 1$) requested.  The seed selection proceeds as follows:
\begin{enumerate}[align=left]
\item[{\bf Stage iii:}] Select Seed Points on the Surface
\begin{enumerate}
\item[{\it Step 7:}] Split the volume $xz$-plane into $V_{g} \times V_{g}$ grid cells.
\item[{\it Step 8:}] Project each point in $H$, onto the $xz$-plane and find the cell it falls in.
\item[{\it Step 9:}] Project the camera location on the $xz$-plane and order the cells in increasing distance of their center to the projected camera point.
\item[{\it Step 10:}] For each grid cell in the order of increasing distance from the camera, randomly select new seed points from $H$, until at most $n_g$ seeds are associated to the cell, while at most $n_s$ seeds are selected in total.
\end{enumerate}
\end{enumerate}

 \subsection{Neighborhood Searching}
 \label{Sec:map_neighborhood}
After the selection of seed points, their neighborhood in the
original point cloud $C$ is found, to provide the local
point data to which new patches will be fit.  Many methods have
been introduced for finding the local 3D points within distance
$r$ (e.g.~foot size) from a seed, for some distance
metric.  We have experimentally validated three search methods
using: (1) K-D trees~\cite{Bentley75}, (2) triangle mesh
datastructures, and (3) back-projections on the image
plane~\cite{RC11}.  In~\cite{KV13}, we evaluate the performance
between the first two in terms of time complexity and their
abilities to handle depth discontinuities; two examples are shown
in Fig.~\ref{Fig:tracking}-(iv).

We decided to use the image plane back-projection method, since it is faster than the other two methods and no extra data-structure is required.  Given the 3D neighborhood-sphere around the seed point and the camera parameters, we can simply backproject it, as a circle in the image plane, centered at the seed's pixel.  The bounding square of pixels that the circle covers, can be easily extracted.  For each one of these $O(r^2)$ pixels, the Euclidean distance of the corresponding 3D points (if any) to the seed point is checked, to see if it is contained to the $r$-sphere.  For a given seed point, this search is linear in the number of checked pixels, which is $O(r^2)$.  The neighborhood finding algorithm proceeds as follows:
\begin{enumerate}[align=left]
  \item[{\bf Stage iv:}] Find Neighborhoods of Seed Points
\begin{enumerate}
\item[{\it Step 11:}] For each seed point $S(i)$, find the $r$-ball neighborhood in the organized point cloud $C$, using the image plane backprojection method.
\item[{\it Step 12:}] For each neighborhood, keep $n_f$ uniformly distributed points.  The parameter $n_f$ is selected depending on the processor speed and the required real-time performance (e.g.~target number of patches to fit per second).
\end{enumerate}
\end{enumerate}

 \subsection{Patch Modeling and Fitting}
 \label{Sec:map_patch_model_fit}
For each point cloud neighborhood found around the seed points, we proceed with patch fitting and validation, as described in Sec.~\ref{Sec:patches}.  After the patch fitting, the principle curvature $\kappa_{\mathrm{min}}\defeq\min(\kappa_x,\kappa_y)$ can be used in a few different ways, depending on the shape of the robot foot sole.  For instance, for a flat footed robot, concave regions, with more than slightly positive $\kappa_{\mathrm{max}}$, could be considered less salient, because the foot cannot fully fit there.  A robot with spherical foot sole, might prefer areas that are not too convex (as the foot would only make contact at a tangent point), but also not too concave to fit.  Fig.~\ref{Fig:tracking}-(v) illustrates $100$ patches that were randomly fitted to salient seed points.  The patch fitting and validation proceeds as follows:
\begin{enumerate}[align=left]
\item[{\bf Stage v:}] Patch Fitting \& Validation
\begin{enumerate}
\item[{\it Step 13:}] Fit a patch $P(i)$ to each neighborhood (Sec.~\ref{Sec:patch_fitting}).
    
\item[{\it Step 14:}] Discard patches with Euclidean residual greater than $d_{\mathrm{max}}$ (Sec.~\ref{Sec:patch_residual}).
    
\item[{\it Step 15:}] Discard patches with areas not sufficiently supported by data, after applying the coverage evaluation (Sec.~\ref{Sec:patch_coverage}).

\item[{\it Step 16:}] Discard patches with min principal curvature less than $\kappa_{\mathrm{min}}$ or max principal curvature greater than $\kappa_{\mathrm{max}}$ (Sec.~\ref{Sec:curvature_validation}).  This step could be adjusted depending on the application.
\end{enumerate}
\end{enumerate}

We estimated the following parameter values from observations of
human selected patches~\cite{KV14}: $\kappa_{\mathrm{min}}=-13.6~\mathrm{m}^{-1}$, $\kappa_{\mathrm{max}}=19.7~\mathrm{m}^{-1}$,
$d_{\mathrm{max}}=0.01~\mathrm{m}$.

 \subsection*{Termination Criteria}
 \label{Sec:map_termination}
For each data frame, one can keep updating the map with new patches until a time or number-of-patches limit (or any other task-specific criteria) has been reached.  Usually we specify initially a desired fraction $\nu$, of the total sampled surface area $S$ that should probabilistically be covered by patches.  If $\nu < 1$, we sparsely sample the surface, while if $\nu > 1$, we oversample it.  For instance, for an $r$-ball neighborhood search and ellipse-bounded paraboloid patch fitting we can estimate the expected number of patches for this criteria as
\begin{equation}\label{Eq:expnumpatches}
  \nu\frac{S}{\pi r^2}.
\end{equation}
When the robot is moving, we also require real-time performance and so impose a time limit on fitting new patches at each map update.

\subsection{Homogeneous Patch Map} \label{Sec:spatial_map}
\begin{figure}
\includegraphics[width=0.5\textwidth]{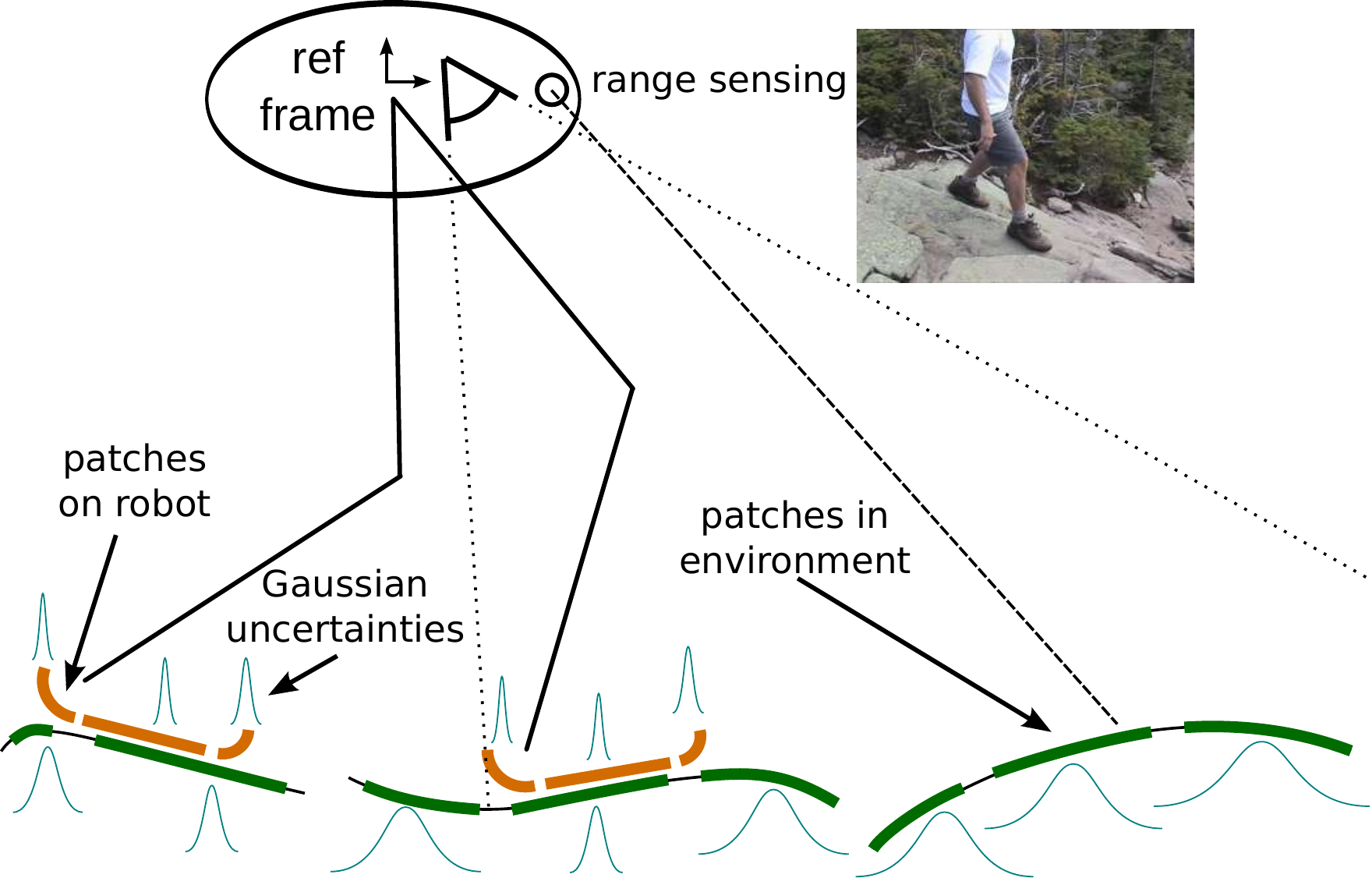}
\caption{The concept of the homogeneous patch map with a sparse set of patches both on the trail surface (green) and possibly also on the robot itself (brown), spatially mapped with quantified uncertainty. \label{Fig:patch-map}}
\end{figure}

Salient patches, from the algorithm proposed in this section, could form the basis for a \emph{homogeneous patch map}; a dynamically maintained local spatial map of curved surface patches, suitable for contacts, both on and around the robot.  Fig.~\ref{Fig:patch-map} illustrates the idea, including both environment surfaces and contact pads on the robot itself (potentially uncertain, due to kinematic errors).  We leave the modeling of patches on the robot, using kinematics and/or proprioception, as future work.  The homogeneous patch map could provide a sparse ``summary'' of relevant contact surfaces for higher-level reasoning.  Upcoming terrain patches can be detected from a distance, using exteroceptive sensing (with high uncertainty), while the map could be further refined, using for instance proprioception for less uncertain patches on the robot itself (e.g.~heel, toe, foot sole), as the robot makes contact with the patches. 

We envision classic elements of SLAM~\cite{SC86}, to be applied
to the introduced map.  For instance, different observations of
the same surface patch can be associated, using propagation of
spatial uncertainty through kinematic chains.  Optimal data
fusion, by Kalman update, is also supported by the patch
covariance matrices.  First-order propagation of uncertainty
through a chain of transforms, with $6\!\times\!6$ covariances
$S_j$, is facilitated by the chain Jacobian $J_c$,
described further in~\cite{VK11}:
\begin{equation}
\Sigma_c=J_c\mathcal{S}J_c^T,\ \mathcal{S}\defeq\diag(S_n,\ldots,S_1).
\end{equation}
where $\Sigma_c\!\!\in\!\!\R^{6\times6}$ is the covariance of the pose of a patch at the end of the chain, relative to the base.

\subsection*{Time Complexity}

Stages I--III are $O(|Z|)$, i.e.~linear in the input.  The implementation of Stage IV is $O(n_sr^2)$ (it could be improved to $O(n_sn_f)$, by switching to breadth-first search on a triangle mesh, but we found the constant factors, favor the image backprojection method for neighborhood search, in practice).  The time complexity of Stage V is dominated by $O(n_sn_f^2)$ for Step 13.  Steps 14--16 are $O(n_sn_f)$.  The worst case time complexity for the whole algorithm is thus $O(|Z|+n_sn_f^2)$.

\section{Patch Tracking} \label{Sec:tracking}

 In the previous section, we described the algorithm that creates
a patch map of the environment around the robot.  During
locomotion, \emph{patch tracking} is also required to find and
add patches to the map online, track them as the robot moves
while new frames are acquired, and drop them when they are left
behind. For tracking, the pose $C_t$ of the range sensor,
with respect to the volume frame, is needed at every frame
$t$.  This is a well-studied problem in the context of
Simultaneous Localization and Mapping (SLAM)~\cite{DB06}. For a
walking robot, a main issue is a potential shaking or jerky
camera motion, during locomotion with intermittent contacts.  We
introduce a GPU-based method, for real-time camera tracking,
including shaky and jerky motions, by extending the Moving Volume
KinectFusion system~\cite{RV12,IKHMNKSHFDF11}.  We review this
system in Sec.~\ref{Sec:track_amvkf} and we describe our adjustments
for our walking robot system.

\subsection{Moving volume KinectFusion review} \label{Sec:track_amvkf}
The KinectFusion algorithm was introduced
in~\cite{IKHMNKSHFDF11}, for real-time 3D depth camera tracking
and dense environment mapping, using a GPU.  To describe the
system, we extend the notion of the volume, introduced in
Sec.~\ref{Sec:Volume}, to a Truncated Signed Distance Formula (TSDF)
volume~\cite{CV96}.  A TSDF volume consists of voxels that
represent a portion of the physical world, by the signed distance
from a physical surface and a confidence weight, representing
data reliability.  A point cloud can be produced using either
raycasting~\cite{PSLHS98} or the marching cubes
method~\cite{LC87}.  As new depth images are acquired, two main
processes alternate:
\begin{enumerate}
 \item[(1)] \textbf{Camera Tracking:} the Generalized Iterative
Closest Point algorithm (GICP)~\cite{SHT09} updates the
camera-to-volume transformation $C_t$
  \item[(2)] \textbf{Data Fusion:} the distance and confidence values are updated in all TSDF voxels, given the newly observed depth image
\end{enumerate}
This system is ideal for our purposes, not only because it can
reliably handle shakes during locomotion, but also because
hole-filling and outlier rejection are some of its properties.
The original system had the TSDF volume fixed in the physical
space.  Our Moving Volume KinectFusion system~\cite{RV12}
extended it to a moving TSDF volume, using the intermittent
moving policies, we described in Sec.~\ref{Sec:Volume}.  This is
implemented by remapping (translating and rotating) the volume
when needed, leaving the camera and the cloud fixed, relative to
the physical world, but moving the TSDF volume to coarsely follow
it.  For legged locomotion, we would also like to have: (1) a
task-specific way to set the inputs to the moving volume policies
and (2) a point cloud around and under the robot's feet, not only
where the real camera is facing.  To handle these two
requirements, we modified the original system as follows.

\subsubsection*{Adaptations to moving volume KinectFusion}

We use the gravity vector (from the IMU) to constrain rotation of the volume, so that the volume $\vec{y}$-axis remains vertical.  This adaptation assumes that the robot is locomoting in a standing-like pose.  We also modify the way that raycasted point cloud is computed from the TSDF.  Instead of raycasting from the real-camera viewpoint, where points near the feet of the robot may not be visible, we do it from a virtual birds-eye view, as long as the TSDF volume voxels have already captured some surface information from previous frames.  We define the virtual birds-eye view camera reference frame to be axis-aligned with the TSDF volume frame, but with its $\vec{z}$-axis pointing down (along the volume frame $\vec{y}$-axis).  The center of projection of the virtual camera is at a fixed offset distance $b_o$ above the location of the real camera, and its $b_r$ width and height (in pixels) are set at fixed resolution, e.g.~$b_r = 200$px.  An example of such a virtual camera can be seen in Fig.~\ref{Fig:tracking}-(i), where the point cloud covers the surrounding area around and under the robot.  This allows patches to be fit under and around the feet, using data that was observed in prior frames.

\subsection{Patch Mapping and Tracking Algorithm} \label{Sec:track_algo}
First, we initialize the volume, such that its $\vec{y}$-axis co-aligns with the gravity vector, and it $\vec{z}$-axis co-aligns with the real camera's forward vector.  We set the initial volume location so the initial camera position is in the middle of the volume.  We split the volume's $xz$-plane into $V_g \times V_g$ grid cells, as described in Sec.~\ref{Sec:map_seeds}.  For every new frame $t$, the full patch mapping and tracking algorithm proceeds as follows:
\begin{enumerate}[align=left]
\item[{\bf Stage I:}] Data Acquisition
\begin{enumerate}
\item[{\it Step 1:}] Acquire a new frame of RGB-D and IMU data.
\end{enumerate}
\item[{\bf Stage II:}] Patch Tracking
\begin{enumerate}
      \item[{\it Step 3:}] Update camera pose $C_t$ in the volume frame, from camera tracking.
      
      \item[{\it Step 4:}] Update the TSDF volume voxels with the fused data.

      \item[{\it Step 5:}] Remap (i.e.~rigid body translation and/or rotation) the volume, according to the moving policies.
            
      \item[{\it Step 6:}] If the TSDF volume was remapped:
\begin{itemize}
        \item Update each patch's pose in the volume frame with the same rigid transformation.
        
        \item Discard the patches that have moved outside the volume.  Optionally, also discard patches that are further than $d_{cp}$ behind the camera.
        
        \item Update the association of existing patches to grid cells in the volume frame $\vec{xz}$-plane.
\end{itemize}
\end{enumerate}
\item[{\bf Stage III:}] Patch Mapping
\begin{enumerate}
\item[{\it Step 8:}] Raycast a point cloud $C$ from the birds-eye view (virtual) camera in the TSDF.
\item[{\it Step 9:}] Find, fit, and validate salient patches (Sec.~\ref{Sec:mapping}), until either a clock time limit $t_m$ is reached or the maximum number of patches $n_s$ are in the map. 
\end{enumerate}
\end{enumerate}
Camera tracking is part of the KinectFusion implementation we use.  We have found it to be quite robust, even in the presence of jerky motions.  However, if camera tracking does fail, then the whole patch map is reset and patch mapping and tracking are re-initialized.

\section{Experimental Results} \label{Sec:exp}
\begin{figure*}
\includegraphics[width=\textwidth]{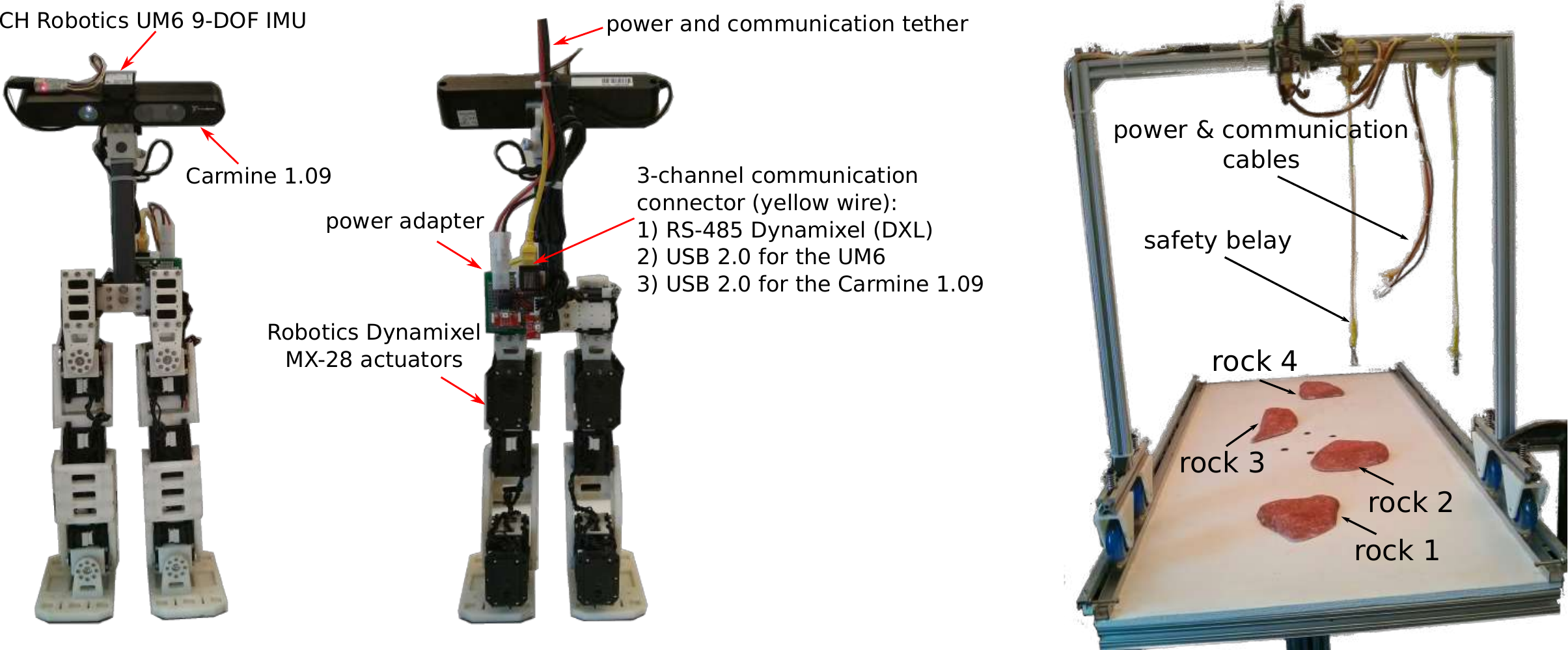}
\caption{Physical hardware of the RPBP mini-biped and the walking table apparatus, with four rocks in fixed positions. \label{Fig:rpbp}}
\end{figure*}
We performed two types of experiments to test the patch mapping and tracking system.  First, we present experiments that test the bio-inspired perceptual framework, using data that were acquired by humans, who were carrying an RGB-D+IMU sensor rig.  We include a quantitative comparison between our algorithm results and patches that humans tend to selected when locomoting.

To experimentally validate the patch mapping and tracking system for robot locomotion, we also developed a 12-DoF mini-biped robotic platform, called the Rapid Prototyped Biped (RPBP), shown in Fig.~\ref{Fig:rpbp}.  We briefly describe its hardware and software design and then present experimental results, where the robot uses our patch mapping and tracking system to identify rocks in a laboratory setup and a simple motion control system to step on them.

 \subsection{Results From Human-Acquired Data}
 \label{Sec:track_exp}
We tested patch fitting and mapping, on data acquired with a hand-held Kinect, from an outdoor trail. Acquisition was on an overcast day, so the sun did not overpower the sensor.

\subsubsection{Patch Mapping Experiment}
 \label{Sec:patch_mapping_exp}
In a first experiment, we applied patch fitting, validation, and mapping, with random seeds on $10$ rock areas, that we selected from the data.  We used random seeds instead of our seed saliency filter, to study how the fitting and validation algorithms behave in isolation.  We let patch mapping run for each dataset, until patch areas covered $90\%$ of the surface, with patch radius $r = 0.1$ m, residual threshold $T_r = 0.01$ m, coverage cell size $w_c = 0.01$ m, coverage threshold factors $\zeta_i = 0.8, \zeta_o = 0.2$, and $T_p = 0.3 A_p/w_c^2$.

Qualitatively (e.g. Fig.~\ref{Fig:patch-fitting}), the fitting and validation algorithm appears to give a good representation of the underline surfaces.   Quantitatively, we measured the following statistics (Table~\ref{tb:exp}): the total and valid number of patches, the number of dropped ones due to residual or coverage, the average Euclidean residual of the valid ones, and the total surface area for each dataset.  We notice, that $79\%$ of the total fitted patches are valid after the evaluation and that the dropped ones, due to residual (depth discontinuities) and coverage (unevenly sample distribution), are equal.  We also notice, that for a 5.43m$^2$ average area, the residual error averages only 4.9~mm, which is a good indicator that the fitting process produces accurate curved patches to represent uneven local surfaces.

\begin{table}
\begin{center}
{\setlength{\tabcolsep}{0.9pt}
	\begin{tabular}{|c||c|c|p{0.5in}|p{0.5in}|c|c|c|}
	\hline
	{\bf Dataset} & \multicolumn{2}{c|}{{\bf Patches}} & \multicolumn{3}{|c|}{{\bf {Dropped patches}}} &	{\bf {Avg res}} & {\bf {Total area}}\\\cline{2-6}
	& {\bf total} & {\bf valid} & {\bf due to \newline residual} & {\bf due to \newline coverage} & {\bf total} &  $(mm)$ & $(m^2)$\\\hline
	1	    & 167 & 142 & 15 & 10 & 25 & 4.6 & 4.57\\
	2	    & 160 & 107 & 18 & 36 & 53 & 5.0 & 3.13\\
	3	    & 231 & 183 & 24 & 24 & 48 & 5.2 & 5.53\\
	4	    & 220 & 164 & 27 & 31 & 56 & 5.0 & 5.01\\
	5	    & 195 & 157 & 17 & 24 & 38 & 5.4 & 4.86\\
	6	    & 235 & 185 & 31 & 20 & 50 & 5.7 & 5.69\\
	7	    & 267 & 213 & 30 & 25 & 54 & 4.4 & 6.54\\
	8	    & 260 & 223 & 16 & 22 & 37 & 4.2 & 7.04\\
	9	    & 187 & 159 & 13 & 16 & 28 & 4.8 & 4.95\\
	10   & 301 & 223 & 30 & 50 & 78 & 5.0 & 6.98\\\cline{1-8}%\hline\hline
	{\bf {avg}}	& \textbf{222} & \textbf{176} & \textbf{22} & \textbf{26} & \textbf{47} & \textbf{4.9} & \textbf{5.43}\\\hline%\hline
	%fake rock	& 65  & 18  & 18 & 44 & 47 & 3.8 & 0.50\\\hline
	\end{tabular}
}
\end{center}
\caption{Quantitative statistics of randomly seeded patch fitting}
\label{tb:exp}
\end{table}

\subsubsection{Human Comparison Experiment}
 \label{Sec:patch_comparison_exp}
As discussed in Sec.~\ref{Sec:map_saliency}, we took video recordings of the footsteps for five humans, walking on trails while carrying an RGB-D+IMU sensor rig.  We visually matched all these footholds to corresponding (pixel, frame) pairs in the RGB-D+IMU data (total 867), where we fit patches (Fig.~\ref{Fig:mapping}).  We then took statistics (min, max, median, average, and standard deviation) of these patches (labeled ``man'' in Table~\ref{Tb:stats}) to set the parameters used in the DoN and DoNG measures.  We ran the patch mapping algorithm to automatically fit salient patches in the same dataset and collected the same statistics (labeled ``auto'' in Table~\ref{Tb:stats}).  Qualitatively, we confirmed that: (1) there are no patches fitted further than 0.7~m from the fixation point, (2) there are no patches in areas with large slope, and (3) there are no patches with very large curvature.  Quantitatively, the statistics for ``auto'' fit patches correspond to the ``man'''s average (human-selected), plus (minus for $\kappa_{\mathrm{min}}$) $3\sigma$.  Across all 82,052 patches that fitted in $832$ data frames, about 1.4\% of patches were dropped due to the curvature, residual, and coverage checks.  This relatively low number, indicates that the saliency checks performed prior to fitting, improved the time efficiency in fitting good patches, compared to the results of Table~\ref{tb:exp}, where patches were seeded randomly.

\begin{table}
\begin{center}
	\begin{tabular}{|r||c|c|c|c|c|c|c|c|c|}
  	\hline
  	 	  {\bf measure} & {\bf min} & {\bf max} & {\bf med} & {\bf avg} & {\bf std} & \\ \hline
	  DoN ($^\circ$) & 0.00 & 26.94 & 4.17 & 4.95 & 3.45 & man\\
                   & 0.00 & 15.31 & 2.83 & 3.65 & 2.85 & auto\\ \hline
	  DoNG ($^\circ$) & 0.24 & 44.85 & 11.37 & 12.34 & 7.54 & man\\
                    & 0.00 & 34.96 &  11.89 & 13.40 & 8.08 & auto\\ \hline
  	$k_{min}$ (m$^{-1}$) & -19.07 & 13.04 & -1.70 & -1.91 & 3.89 & man\\
                         & -11.97 & 7.64 & -0.87 & -1.14 & 1.75 & auto\\ \hline
  	$k_{max}$ (m$^{-1}$) & -7.87 & 28.94 & 3.78 & 4.62 & 5.02 & man\\
                         & -7.81 & 16.97 & 1.01 & 1.32 & 1.76 & auto\\ \hline
	\end{tabular}
\end{center}
\caption{Quantitative statistics of human selected vs. automatically fitted patches.}
\label{Tb:stats}
\end{table}

\subsubsection*{Computational Cost}

On commodity hardware (one 2.50~GHz core, 8GB RAM) and a  {$640~\times~480$} RGB-D image, preprocessing (bilateral filter and downsampling) run in $\sim$20~ms total. Normal computation, DtFP, DoN, and DoNG saliency take $\sim$35~ms.  Neighborhood finding takes $\sim$0.03~ms per seed, and patch fitting and validation are $\sim$0.8~ms total per neighborhood, with $n_f=50$.  The total time elapsed per frame is $20+35+0.83n_p~{\rm ms}$, where $n_p$ is the number of patches actually added.  $n_p$ can range from $0$, in the case that the map is already full (or there are no new seed points), up to $n_g V_g^2$.  In practice we additionally limit the total time spent per frame to e.g.~$100~{\rm ms}$, allowing up to around $50$ patches to be added per frame, in this configuration.

\subsection{{RPBP} Robot and Software Interface} \label{Sec:biped_rpbp}
RPBP is a 3D printed mini-biped robot, which weighs 1.7~kg and it
is $47$cm tall.  Its 6-DoF legs are kinematically similar
to the DARwIn-OP humanoid~\cite{HTA11}.  The actuators are the
Robotics Dynamixel MX-28, with high resolution magnetic rotation
sensors and PID control.  We used the PrimeSense Carmine 1.09
depth camera, which is a sensor similar to the Microsoft Kinect,
but with better close-range capabilities.  We mounted a CH
Robotics UM6 IMU sensor on top of it.  The UM6 has a 3-DoF
accelerometer, a 3-DoF gyroscope, and a 3-DoF magnetometer (9-DoF
in total).  The robot has a lightweight tether for off-board
power and 3 communication channels to an external computer.   We
used RS-485 Dynamixel (DXL) communication for controlling the
motors and USB 2.0 for the UM6 IMU and the Carmine 1.09 camera.

The software interface is implemented in C++ and is divided into
the perception and control subsystems (Fig.~\ref{Fig:overview}).  The
perception subsystem builds on PCL~\cite{RC11} and includes: (1)
a library we wrote that implements an RGB-D+IMU frame grabber,
providing 30fps depth and 100fps IMU data, (2) our Moving Volume
KinectFusion library, and (3) our Surface Patch Library (SPL)
library~\cite{SPL14}, that implements the patch mapping system,
where salient patches fit in the environment.  The perception
system provides a set of patches to the control system, which
includes our Robotis Dynamixel-based communication library and
the RPBP walk controller.  The walk controller maps from a small
library of known patch positions relative to the robot to
corresponding predefined step motions.  At runtime, the walk
control system finds matches between the patches from the
perception system and those in the library, then executes the
corresponding motion sequence, when an approximate match is
found.

\subsection{Application to Bipedal Foot Placement} \label{Sec:biped_exp}
\begin{figure*}
\includegraphics[width=\textwidth]{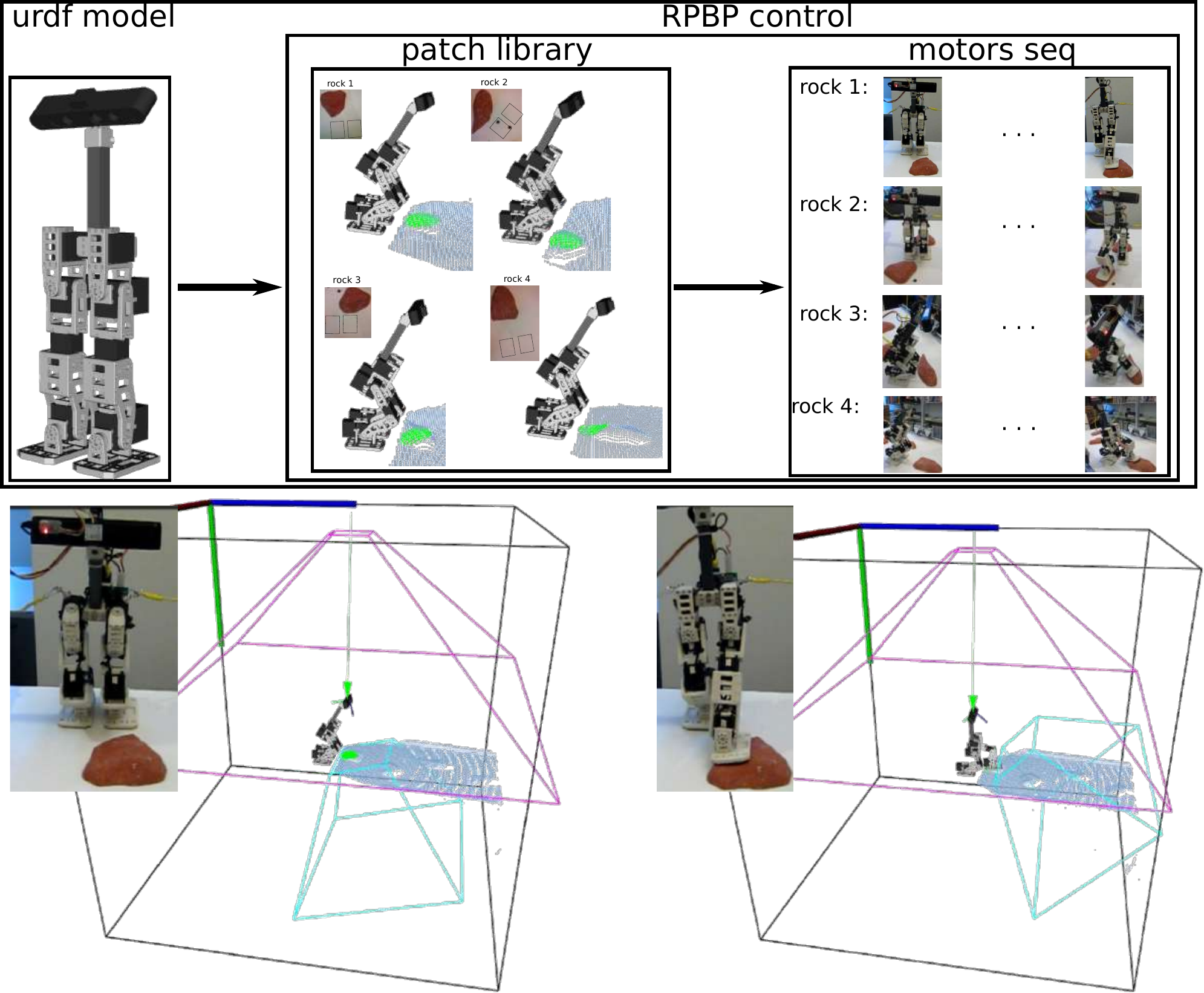}
\caption{{Top:} Software interface for the RPBP robot. {Bottom:} RPBP detects a patch on rock 1 and places its foot, using the approximately matching predefined motion. \label{Fig:biped_exp}}
\end{figure*}

Bipedal locomotion in rough terrain is one of the most
challenging tasks in robotics.  In this paper, we study the case
of non-point feet, where the robot uses the foot sole contact
area to support torques for balance.  In this section, we test
our perception hypothesis, by integrating the proposed patch
mapping and tracking framework into a real-time footfall
selection system on the RPBP biped, for foot placement on some
rocks.  The focus of the experiments is perception and we thus
use a very simple control system, where the robot executes a set
of predefined leg motion primitives~\cite{MLB13}, driven from the
type of patch that is selected for contact.  We thus manually
train a library of patches and a motion sequence for each one of
them.

Our apparatus (Fig.~\ref{Fig:rpbp}), includes a table with 4 solid rocks in fixed positions.  The robot is always attached to a safety belay, which is not intended to affect its motion significantly, i.e.~it is slack.  We manually posed the robot to place its foot on four different types of patches on rocks, recording motion keyframes into a library of stored stepping motions.  We then, placed it in front of one of the same rocks and let it create a patch map and find a match between the trained patches and one of those in the map.  If a match is found, we let it run the corresponding trained motion sequence and place its foot on the rock.  To our knowledge this is the first perception system that lets a biped place its foot on very rough terrain.

\subsubsection*{Foot Placement Training}

During the training, we first let the robot be in its starting \emph{lookdown} pose, as appears in Fig.~\ref{Fig:biped_exp}.  We place the robot in a pre-defined position in front of each of the four rocks and we let our KinectFusion system provide us with a point cloud from the virtual camera.  For each rock, we manually select a neighborhood where we would like the robot to place its foot and we fit a patch.  We then, train the robot to place its foot on the patch, with a corresponding motion sequence.  This forms a library, where patches are stored relative to the robot and corresponding motion sequences.  Similarly, we could train the robot to place its feet on various other positions, but this goes beyond the intention of this experiment, that focuses on perception, not control or trajectory planning.

\subsubsection*{Foot Placement Tests}

We place RPBP in front of each rock in the lookdown pose.  Then, we let the perception system create a patch map using seeds in the area around the robot's feet, i.e.~$10$cm in front of the robot.  A patch matching is then performed, by comparing every patch in the map with every trained one.  The similarity comparison between two patches proceeds as follows:  
\begin{enumerate}
  \item[(1)] check if the patches are of the same surface type (flat or elliptic/hyperbolic/cylindric paraboloid)
  \item[(2)] check if the absolute difference between their boundary parameters are smaller than a threshold $d_s = 0.015$ m
  \item[(3)] check if the absolute difference between the curvatures are smaller than a threshold $k_s=5$m$^{-1}$
  \item[(4)] check if the angle between their normal vectors ($\vec{z}_{\ell}$ axis) is smaller than a threshold $a_s = 20^{\circ}$ (we also consider discrete rotations of the patches here, to check all symmetry cases)
  \item[(5)] check if the distance between their position relative to the robot is smaller than a threshold $r_s = 0.01$ m
\end{enumerate}
To compare normal vector angles, we consider every possible symmetry.  If any patch in the map matches with a trained one, we execute the corresponding motion sequence.

We ran the aforementioned experiment twenty times for each rock,
in which the robot never failed to match the correct trained
patch and successfully run the motion sequence of placing its
foot on the rock.  Success was defined as maintaining balance and
ending with the foot on the rock.  We also placed the robot in
front of a few other rocks, that were not in the training set,
and only when the geometry of the rock was matching to the
trained one (in terms of patches), the motion was executed.  An
example is visualized in Fig.~\ref{Fig:biped_exp}, where the robot
detects a match with the corresponding trained patch and executes
the motion sequence.  During the motion execution, we also
noticed that the robot tracks very accurately a matched patch.
More experimental details can be found
in~\cite{SPL14,Kanoulas14}.

Our basic controller is not a full solution to rough-terrain bipedal locomotion, so this experiment did not fully test the ability of the robot to walk continuously, while mapping and tracking patches. Thus, we performed a separate experiment, where RPBP walked in a circle on a flat surface, while mapping and tracking patches on nearby rocks.  This verified that the KinectFusion camera tracking was sufficient to maintain camera tracking on a walking robot and that our patch mapping and tracking algorithms were able to maintain a patch map around the robot as it moves through the environment.  A video with a set of patch mapping and tracking results can be reached at the following link: \url{https://sites.google.com/view/cpmat}

The patch mapping framework has been also successfully used on the WALK-MAN full-size humanoid robot during the DARPA Robotics Challenge (DRC) 2015 finals.  In particular, we developed an interface, where the pilot could select manually a sequence of areas to fit foothold patches in the environment and use the roughly flat ones for dynamic stepping.

 \section{Discussion and Limitations}

\subsection{Discussion}

There are various ways to represent environment surfaces in the
computational stack from perception to planning and control.  A
common approach is to allow the perception system to report
arbitrarily complex dense surface geometries -- e.g.~point clouds,
triangle meshes, NURBS patches -- to the planning and control
systems.  Our set of curved patch types is a proposal in the
opposite direction where the perception system identifies a
sparse set of (possibly overlapping) foot-scale patches of nearby
terrain, each of which is described by only a few parameters.
This is an architectural choice intended to make building the
planning and control algorithms more tractable.  The planning
system could select a specific patch for the next footfall and
communicate just that patch to the control system.  The dynamics
of the support phase could then be handled by considering contact
with only that patch.  In this paper we focus mostly only on
perception, but in~\cite{KZNKCT2017,KWRZTSCT2018} we implemented
fast footstep planning using the curved patches introduced here.

We include curved vs just flat patches for several reasons.  First, in very rough terrain, it may not always be feasible to find flat surfaces.  Humans do sometimes step on rocks with major curvatures, such that either tangential/rolling (positive curvature) or fixed (negative curvature) contacts take place.  Bipedal robots which eventually approach the level of capability to walk on very rough terrain that humans do may also need to utilize such non-flat contacts.

Second, though it is true that most current humanoid and bipedal
robots have flat feet, in the future this may not always be the
case.  Using curved patches also enables extension to quadrupeds
which often have spherical feet.  Curved patches can also be used
for other tasks, such as grasp affordances for
manipulation~\cite{KLCT2016}.

Third, curved patches could also be used to represent surfaces on
the robot itself, as shown in Fig.~\ref{Fig:patch-map}, so that
robot-environment interactions could be reasoned about as
homogeneous patch--patch contacts~\cite{KZNKCT2017}.

One could imagine a different taxonomy of patch types than the one we proposed.  For instance, the four paraboloid types with different curvature signs could be reduced to a single generic one.  Our reasoning to keep them separate is that contact kinematics would likely be significantly different for these types, so identifying them in the perception system could offer a clearer choice of options to the planning and control systems.  Similarly, the four planar types with different boundary shapes could be reduced to one with a generic boundary, or the planar type could just be a special case of a generic paraboloid patch which happens to have zero curvatures.  We feel that these choices would mostly just push the task of differentiating these types of surfaces from the perception to the planning and control systems.  (We mainly include spherical and cylindrical patches in the taxonomy to more accurately represent man-made surfaces, but they could be cleanly omitted from implementations if this is not needed.)

\subsection{Limitations}

The patch mapping and tracking framework we propose shows some positive and promising results on the difficult problem of bipedal foot placement on non-flat surfaces.  The proposed system has some limitation though that we discuss briefly in this section.

We use RGB-D data from a range sensor, e.g.~a Kinect or
PrimeSense Carmine 1.09.  These sensors are often very accurate
for data acquired close to the device.  Thus, the data
uncertainty, especially for a mini-bipedal robot, is smaller than
the uncertainty that is introduced, for instance, due to the
kinematics of the legs.  We have done an uncertainty analysis
in~\cite{KTV2016}, but integrating uncertainty from the robot
kinematics into the patch map would be an interesting extension
we have not yet developed.

Using the patch uncertainty based only on RGB-D sensor uncertainty was thus not strictly necessary for a mini-biped.  This is not true when more noisy sensors on bigger humanoid robots are used, such as the Multisense-SL sensor that is installed on our WALK-MAN humanoid.  In this case, the sensor uncertainty can play an important role.

Moreover, we use only the depth information of the range sensors.  The RGB data are ignored, but could play an important role, for instance when semantics about the environment need to be extracted, e.g.~for finding rigid terrains.  In addition, from the IMU data we considered only the gravity information.  Other data could be extracted and used in patch selection, such as the direction/speed of movement.

Our curved patch models use second degree polynomials which may be concave, convex, flat, or saddle-shaped.  A limitation comes with environment surfaces that have more complex structure, but we believe that this can be handled with impedance or torque control stepping methods, in a low-level control fashion.  This is also related to the patch-size.  In our current work, we pick patches slightly bigger than the foot size.  In nature, though, there are cases a foot may need to make partial contact with smaller surfaces.  Also related is the patch validation that we introduced, which drops patches that are partially not appropriate for contact; it may be interesting to attempt to use these patches if possible.

A limitation in the patch mapping and tracking part is the use of the GPU for the Moving Volume KinectFusion system.  Even though most of humanoids at the moment are able to carry GPUs, it would be interesting to make the approach more energy efficient by using the CPU instead.

Given that graph-based path planners may need a bigger set of patches to expand on their states, it would be interesting to make the patch fitting faster and/or design a real-time path planner-oriented seed selection algorithm.  In terms of the human-based saliency measures, an interesting direction could be the analysis of human gaze patterns.

Last but not least, the experimental part tests a simple
controller to prove the concept of the patch-based contact
stepping.  A better quasi-static or dynamic controller integrated
with a footstep path planner is required for more meaningful and
realistic locomotion experiments.  In~\cite{KZNKCT2017}, we used
a dynamic locomotion module.  In addition, more
rough/unstructured rocks should be used in the future
experiments. 

\section{Conclusion}

This paper introduced a novel real-time mapping and tracking system for bipedal locomotion based on bounded curved patches.  We ran experiments for mapping and tracking patches and for foot placement on some rocky terrains using a mini-biped with on-board RGB-D and IMU sensors.  In separate work, our patch mapping and tracking algorithms were also used for fitting manually picked footholds, during the 2015 DARPA Robotics Challenge, for the WALK-MAN humanoid robot.  We envision this perception framework to be part of a path planning and humanoid walking system, using the relatively sparse map of curved patches, instead of a dense environment representation.  We also plan to use the patch uncertainty in a state estimator system data fusion and foothold selection.
\ifCLASSOPTIONcaptionsoff
  \newpage
\fi

\bibliographystyle{IEEEtran}
\bibliography{IEEEabrv,ras_2017.bib}

\begin{thebibliography}{10}
\providecommand{\url}[1]{#1}
\csname url@rmstyle\endcsname
\providecommand{\newblock}{\relax}
\providecommand{\bibinfo}[2]{#2}
\providecommand\BIBentrySTDinterwordspacing{\spaceskip=0pt\relax}
\providecommand\BIBentryALTinterwordstretchfactor{4}
\providecommand\BIBentryALTinterwordspacing{\spaceskip=\fontdimen2\font plus
\BIBentryALTinterwordstretchfactor\fontdimen3\font minus
  \fontdimen4\font\relax}
\providecommand\BIBforeignlanguage[2]{{%
\expandafter\ifx\csname l@#1\endcsname\relax
\typeout{** WARNING: IEEEtran.bst: No hyphenation pattern has been}%
\typeout{** loaded for the language `#1'. Using the pattern for}%
\typeout{** the default language instead.}%
\else
\language=\csname l@#1\endcsname
\fi
#2}}

\bibitem{KS09}
S.~Kajita and T.~Sugihara, ``{Humanoid Robots in the Future},'' \emph{Advanced
  Robotics}, vol.~23, pp. 1527--1531, 2009.

\bibitem{CLCKHK05}
J.~Chestnutt, M.~Lau, G.~Cheung, J.~Kuffner, J.~Hodgins, and T.~Kanade,
  ``{Footstep Planning for the Honda {ASIMO} Humanoid},'' in \emph{{IEEE} Int.
  Conf. on Rob. and Aut. (ICRA)}, 2005, pp. 629--634.

\bibitem{MCKK2005}
P.~Michel, J.~Chestnutt, J.~Kuffner, and T.~Kanade, ``{Vision-guided Humanoid
  Footstep Planning for Dynamic Environments},'' in \emph{IEEE-RAS
  International Conference on Humanoid Robots (Humanoids)}, 2005, pp. 13--18.

\bibitem{NCK12}
K.~Nishiwaki, J.~E. Chestnutt, and S.~Kagami, ``{Autonomous Navigation of a
  Humanoid Robot over Unknown Rough Terrain using a Laser Range Sensor},''
  \emph{International Journal of Robotic Research}, vol.~31, no.~11, pp.
  1251--1262, 2012.

\bibitem{RBNP08}
M.~Raibert, K.~Blankespoor, G.~Nelson, R.~Playter, and {the BigDog Team},
  ``{BigDog, The Rough-Terrain Quadruped Robot},'' \emph{17th World Cong. of
  the Int. Fed. of Aut. Control}, pp. 10\,822--10\,825, 2008.

\bibitem{Wiedebach2016}
G.~Wiedebach \emph{et~al.}, ``{Walking on Partial Footholds Including Line
  Contacts with the Humanoid Robot ATLAS},'' in \emph{IEEE-RAS 16th Int. Conf.
  on Humanoid Robots (Humanoids)}, 2016, pp. 1312--1319.

\bibitem{MLB13}
D.~Maier, C.~Lutz, and M.~Bennewitz, ``{Integrated Perception, Mapping, and
  Footstep Planning for Humanoid Navigation Among 3D Obstacles},'' in
  \emph{{IEEE/RSJ} International Conference on Intelligent Robots and Systems
  {IROS}}, 2013.

\bibitem{FMDWAMT15}
M.~F. Fallon, P.~Marion, R.~Deits, T.~Whelan, M.~Antone, J.~McDonald, and
  R.~Tedrake, ``{Continuous Humanoid Locomotion over Uneven Terrain using
  Stereo Fusion},'' in \emph{15th IEEE-RAS International Conference on Humanoid
  Robots (Humanoids)}, 2015.

\bibitem{VK11}
M.~Vona and D.~Kanoulas, ``{Curved Surface Contact Patches with Quantified
  Uncertainty},'' in \emph{IEEE/RSJ Int. Conf. on Intelligent Robots and
  Systems (IROS)}, 2011, pp. 1439 --1446.

\bibitem{KV13}
D.~Kanoulas and M.~Vona, ``{Sparse Surface Modeling with Curved Patches},'' in
  \emph{IEEE International Conference on Robotics and Automation (ICRA)}, 2013,
  pp. 4209--4215.

\bibitem{KV14}
------, ``{Bio-Inspired Rough Terrain Contact Patch Perception},'' in
  \emph{IEEE Int. Conf. on Rob. and Autom. (ICRA)}, 2014, pp. 1719--1724.

\bibitem{SPL14}
------, ``{Surface Patch Library (SPL)},'' in \emph{{IEEE Int.Conf.on Rob. and
  Autom. (ICRA) Workshop: MATLAB/Simulink for Robotics Education and
  Research}}, 2014, \url{http://dkanou.github.io/projects/spl/}.

\bibitem{DRC-what-happened}
DRC-Teams, ``{What Happened at the DARPA Robotics Challenge?}'' 2015,
  \url{www.cs.cmu.edu/~cga/drc/events}.

\bibitem{GHB11}
J.~Garimort, A.~Hornung, and M.~Bennewitz, ``{Humanoid Navigation with Dynamic
  Footstep Plans},'' in \emph{IEEE International Conference on Robotics and
  Automation (ICRA)}, 2011, pp. 3982--3987.

\bibitem{HDLB12}
A.~Hornung, A.~Dornbush, M.~Likhachev, and M.~Bennewitz, ``{Anytime
  Search-Based Footstep Planning with Suboptimality Bounds},'' in
  \emph{IEEE-RAS Int. Conf. on Humanoid Robots (Humanoids)}, 2012.

\bibitem{KB16}
P.~Karkowski and M.~Bennewitz, ``{Real-time Footstep Planning using a Geometric
  Approach},'' in \emph{IEEE International Conference on Robotics and
  Automation (ICRA)}, May 2016, pp. 1782--1787.

\bibitem{OII03}
K.~Okada, M.~Inaba, and H.~Inoue, ``{Walking Navigation System of Humanoid
  Robot Using Stereo Vision Based Floor Recognition and Path Planning with
  Multi-Layered Body Image},'' in \emph{IEEE/RSJ Int. Conf. on Intel. Robots
  and Systems (IROS)}, 2003, pp. 2155--2160.

\bibitem{OOHI05}
K.~Okada, T.~Ogura, A.~Haneda, and M.~Inaba, ``{Autonomous 3D Walking System
  for a Humanoid Robot Based on Visual Step Recognition and 3D Foot Step
  Planner},'' in \emph{Robotics and Automation, 2005. ICRA 2005. Proceedings of
  the 2005 IEEE International Conference on}, April 2005, pp. 623--628.

\bibitem{GFF08}
J.-S. Gutmann, M.~Fukuchi, and M.~Fujita, ``{3D Perception and Environment Map
  Generation for Humanoid Robot Navigation},'' \emph{Int. J. of Robotics
  Research}, vol.~27, no.~10, pp. 1117--1134, 2008.

\bibitem{CTSNKK09}
J.~Chestnutt, Y.~Takaoka, K.~Suga, K.~Nishiwaki, J.~Kuffner, and S.~Kagami,
  ``{Biped Navigation in Rough Environments Using On-board Sensing},'' in
  \emph{Proceedings of the 2009 IEEE/RSJ International Conference on
  Intelligent Robots and Systems}, ser. IROS'09.\hskip 1em plus 0.5em minus
  0.4em\relax Piscataway, NJ, USA: IEEE Press, 2009, pp. 3543--3548.

\bibitem{UNOI2014}
R.~Ueda, S.~Nozawa, K.~Okada, and M.~Inaba, ``{Biped Humanoid Navigation System
  Supervised through Interruptible User-Interface with Asynchronous Vision and
  Foot Sensor Monitoring},'' in \emph{IEEE-RAS Int. Conf. on Humanoid Robots
  (Humanoids)}, 2014, pp. 273--278.

\bibitem{BVKEK13}
S.~Brossette, J.~Vaillant, F.~Keith, A.~Escande, and A.~Kheddar, ``{Point-Cloud
  Multi-Contact Planning for Humanoids: Preliminary Results},'' in
  \emph{Cybarnetics and Intelligent Systems Robotics, Automation and
  Mechatronics (CISRAM)}, vol.~1, 2013, p.~6.

\bibitem{RGMSHS13}
O.~E. Ramos, M.~Garc\'ia, N.~Mansard, O.~Stasse, J.-B. Hayet, and P.~Sou\`eres,
  ``{Towards Reactive Vision-Guided Walking on Rough Terrain: An
  Inverse-Dynamics Based Approach},'' in \emph{Workshop on Visual Navigation
  for Humanoid Robots (IEEE ICRA)}, 2013.

\bibitem{KMK13}
N.~Kita, M.~Morisawa, and F.~Kanehiro, ``{Foot Landing State Estimation from
  Point Cloud at a Landing Place},'' in \emph{13th IEEE-RAS Humanoids}, 2013,
  pp. 252--259.

\bibitem{NIHMKDKSHF11}
R.~A. Newcombe \emph{et~al.}, ``{KinectFusion: Real-Time Dense Surface Mapping
  and Tracking},'' in \emph{{IEEE} International Symposium on Mixed and
  Augmented Reality (ISMAR)}, 2011.

\bibitem{DT2014}
R.~Deits and R.~Tedrake, ``{Footstep Planning on Uneven Terrain with
  Mixed-Integer Convex Optimization},'' in \emph{IEEE-RAS Int. Conference on
  Humanoid Robots (Humanoids)}, 2014, pp. 279--286.

\bibitem{Kohlbrecher14}
S.~Kohlbrecher, A.~Romay, E.~Stumpf, A.~Gupta, O.~V. Stryk, F.~Bacim, D.~A.
  Bowman, A.~Goins, R.~Balasubramanian, and D.~C. Conner, ``{Human-Robot
  Teaming for Rescue Missions: Team ViGIR’s Approach to the 2013 DARPA
  Robotics Challenge Trials},'' \emph{Journal of Field Robotics (JFR)}, 2014.

\bibitem{SKSC2016}
A.~Stumpf, S.~Kohlbrecher, O.~von Stryk, and D.~C. Conner, ``{Open Source
  Integrated 3D Footstep Planning Framework for Humanoid Robots},'' in
  \emph{IEEE Int. Conf. on Humanoid Rob. (Humanoids)}, 2016.

\bibitem{KOB2016}
P.~Karkowski, S.~O{\ss}wald, and M.~Bennewitz, ``{Real-time footstep planning
  in 3D environments},'' in \emph{IEEE-RAS 16th International Conference on
  Humanoid Robots (Humanoids)}, 2016, pp. 69--74.

\bibitem{SKCS14}
A.~Stumpf, S.~Kohlbrecher, D.~Conner, and O.~von Stryk, ``{Supervised Footstep
  Planning for Humanoid Robots in Rough Terrain Tasks using a Black Box Walking
  Controller},'' in \emph{IEEE-RAS Int. Conference on Humanoid Robots
  (Humanoids)}, 2014, pp. 287--294.

\bibitem{KC14}
O.~Khatib and S.-Y. Chung, ``{SupraPeds: Humanoid Contact-Supported Locomotion
  for 3D Unstructured Environments},'' in \emph{Proceedings of the IEEE
  International Conference on Robotics and Automation (ICRA)}, 2014.

\bibitem{WMBHRR10}
D.~Wooden, M.~Malchano, K.~Blankespoor, A.~Howard, A.~A. Rizzi, and M.~Raibert,
  ``{Autonomous Navigation for BigDog},'' in \emph{{IEEE} Int. Conf. on Rob.
  and Aut. (ICRA)}, 2010, pp. 4736--4741.

\bibitem{SHG12}
A.~Stelzer, H.~Hirschm{\"u}ller, and M.~G{\"o}rner, ``{Stereo-Vision-Based
  Navigation of a Six-Legged Walking Robot in Unknown Rough Terrain},''
  \emph{Int. J. Robotics Research}, vol.~31, no.~4, pp. 381--402, 2012.

\bibitem{PMPKRB09}
C.~Plagemann, S.~Mischke, S.~Prentice, K.~Kersting, N.~Roy, and W.~Burgard,
  ``{A Bayesian Regression Approach to Terrain Mapping and an Application to
  Legged Robot Locomotion},'' \emph{Journal of Field Robotics}, vol.~26, pp.
  789--811, 2009.

\bibitem{KKN09}
J.~Z. Kolter, Y.~Kim, and A.~Y. Ng, ``{Stereo Vision and Terrain Modeling for
  Quadruped Robots},'' in \emph{IEEE International Conference on Robotics and
  Automation (ICRA)}, 2009, pp. 3894--3901.

\bibitem{KBPMS10}
M.~Kalakrishnan, J.~Buchli, P.~Pastor, M.~Mistry, and S.~Schaal, ``{Learning,
  Planning, and Control for Quadruped Locomotion over Challenging Terrain},''
  \emph{Int. J. of Robotics Research}, no.~2, pp. 236--258, 2010.

\bibitem{BPP10}
D.~Belter, P.~{\L}abȩcki, and P.~Skrzypczynski, ``{Map-based Adaptive
  Foothold Planning for Unstructured Terrain Walking},'' in \emph{IEEE Int.
  Conf. on Robotics and Automation (ICRA)}, 2010, pp. 5256--5261.

\bibitem{BS12}
D.~Belter and P.~Skrzypczynski, ``{Precise Self-Localization of a Walking Robot
  on Rough Terrain Using PTAM},'' in \emph{Adaptive Mobile Robotics}.\hskip 1em
  plus 0.5em minus 0.4em\relax World Scientific, 2012, pp. 89--96.

\bibitem{WFDHKS15}
M.~Wermelinger, P.~Fankhauser, R.~Diethelm, P.~Krusi, R.~Siegwart, and
  M.~Hutter, ``{Navigation Planning for Legged Robots in Challenging
  Terrain},'' in \emph{IEEE/RSJ IROS}, 2016, pp. 1184--1189.

\bibitem{MHWCS15}
C.~Mastalli, I.~Havoutis, A.~W. Winkler, D.~G. Caldwell, and C.~Semini,
  ``{On-line and On-board Planning and Perception for Quadrupedal
  Locomotion},'' in \emph{IEEE TePRA}, 2015.

\bibitem{KB2017}
T.~Klamt and S.~Behnke, ``{Anytime Hybrid Driving-Stepping Locomotion
  Planning},'' in \emph{IEEE/RSJ Int. Conf. on Intelligent Robots and Systems
  (IROS)}, 2017.

\bibitem{CPP99}
C.-M. Chew, J.~Pratt, and G.~Pratt, ``{Blind Walking of a Planar Bipedal Robot
  on Sloped Terrain},'' in \emph{IEEE Int. Conf. on Robotics and Automation
  (ICRA)}, 1999, pp. 381--386.

\bibitem{MHB12}
D.~Maier, A.~Hornung, and M.~Bennewitz, ``{Real-Time Navigation in 3D
  Environments Based on Depth Camera Data},'' in \emph{IEEE-RAS International
  Conference on Humanoid Robots (HUMANOIDS)}, 2012.

\bibitem{FAF86}
O.~Faugeras, N.~Ayache, and B.~Faverjon, ``{Building Visual Maps by Combining
  Noisy Stereo Measurements},'' in \emph{IEEE Int. Conf. on Robotics and
  Automation (ICRA).}, 1986, pp. 1433--1438.

\bibitem{MS87}
L.~Matthies and S.~A. Shafer, ``\BIBforeignlanguage{English}{{Error Modeling in
  Stereo Navigation}},'' pp. 135--144, 1990.

\bibitem{KE12}
K.~Khoshelham and S.~O. Elberink, ``{Accuracy and Resolution of Kinect Depth
  Data for Indoor Mapping Applications},'' in \emph{Sensors}, vol.~12, 2012,
  pp. 1437--1454.

\bibitem{DVX13}
I.~Dryanovski, R.~G. Valenti, and J.~Xiao, ``{Fast Visual Odometry and Mapping
  from RGB-D Data},'' in \emph{IEEE International Conference on Robotics and
  Automation (ICRA)}, May 2013, pp. 2305--2310.

\bibitem{HBJFBGBEFF96}
A.~Hoover, G.~Jean-Baptiste, X.~Jiang, P.~J. Flynn, H.~Bunke, D.~Goldgof,
  K.~Bowyer, D.~Eggert, A.~Fitzgibbon, and R.~Fisher, ``{An Experimental
  Comparison of Range Image Segmentation Algorithms},'' \emph{IEEE Transactions
  on Pattern Analysis and Machine Intelligence}, vol.~18, no.~7, pp. 673--689,
  1996.

\bibitem{PBJB98}
M.~W. Powell, K.~W. Bowyer, X.~Jiang, and H.~Bunke, ``{Comparing Curved-Surface
  Range Image Segmenters},'' in \emph{Int. Conf. on Computer Vision (ICCV)},
  1998.

\bibitem{WTHNY01}
C.~Wang, H.~Tanahashi, H.~Hirayu, Y.~Niwa, and K.~Yamamoto, ``Comparison of
  local plane fitting methods for range data,'' in \emph{Computer Vision and
  Pattern Recognition (CVPR)}, 2001, pp. 663--669.

\bibitem{Kanatani05}
K.~Kanatani, \emph{{Statistical Optimization for Geometric Computation: Theory
  and Practice}}.\hskip 1em plus 0.5em minus 0.4em\relax Dover Publications,
  Incorporated, 2005.

\bibitem{PVB09}
K.~Pathak, N.~Vaskevicius, and A.~Birk, ``{Revisiting Uncertainty Analysis for
  Optimum Planes Extracted from 3D Range Sensor Point-Clouds},'' in \emph{IEEE
  International Conference on Robotics and Automation (ICRA)}, 2009, pp.
  1631--1636.

\bibitem{Petitjean02}
S.~Petitjean, ``{A Survey of Methods for Recovering Quadrics in Triangle
  Meshes},'' \emph{ACM Comp. Surv.}, vol.~34, pp. 211--262, 2002.

\bibitem{DNC07}
M.~Dai, T.~S. Newman, and C.~Cao, ``{Least-Squares-Based Fitting of
  Paraboloids},'' \emph{Pattern Recognition}, vol.~40, pp. 504--515, 2007.

\bibitem{MDR04}
N.~D. Molton, A.~J. Davison, and I.~D. Reid, ``{Locally Planar Patch Features
  for Real-Time Structure from Motion},'' in \emph{British Machine Vision
  Conference (BMVC)}, Sep 2004.

\bibitem{RV12}
H.~Roth and M.~Vona, ``{Moving Volume KinectFusion},'' in \emph{British Machine
  Vision Conference}, 2012.

\bibitem{Everett95}
H.~R. Everett, \emph{{Sensors for Mobile Robots: Theory and
  Application}}.\hskip 1em plus 0.5em minus 0.4em\relax Natick, MA, USA: A. K.
  Peters, Ltd., 1995.

\bibitem{SNS11}
R.~Siegwart, I.~R. Nourbakhsh, and D.~Scaramuzza, \emph{Introduction to
  Autonomous Mobile Robots}, 2nd~ed.\hskip 1em plus 0.5em minus 0.4em\relax The
  MIT Press, 2011.

\bibitem{RC11}
R.~B. Rusu and S.~Cousins, ``{3D is here: Point Cloud Library (PCL)},'' in
  \emph{IEEE Int. Conf. on Robotics and Automation (ICRA)}, 2011.

\bibitem{SS02}
D.~Scharstein and R.~Szeliski, ``{A Taxonomy and Evaluation of Dense Two-Frame
  Stereo Correspondence Algorithms},'' \emph{Int. J. of Computer Vision},
  vol.~47, no. 1-3, pp. 7--42, 2002.

\bibitem{ML05}
D.~Murray and J.~J. Little, ``{Patchlets: Representing Stereo Vision Data with
  Surface Elements},'' in \emph{7th IEEE Workshops on Application of Computer
  Vision (WACV/MOTION)}, 2005, pp. 192--199.

\bibitem{KM10}
K.~Konolige and P.~Mihelich, ``Technical description of kinect calibration,''
  2010, \url{www.ros.org/wiki/kinect_calibration/technical}.

\bibitem{ELF97}
D.~Eggert, A.~Lorusso, and R.~Fisher,
  ``\BIBforeignlanguage{English}{{Estimating 3-D Rigid Body Transformations: a
  Comparison of Four Major Algorithms}},''
  \emph{\BIBforeignlanguage{English}{Machine Vision and Applications}}, vol.~9,
  no. 5--6, pp. 272--290, 1997.

\bibitem{Srinivasan03}
V.~Srinivasan, \emph{{Theory of Dimensioning}}.\hskip 1em plus 0.5em minus
  0.4em\relax Marcell Dekker, 2003.

\bibitem{Gra98}
F.~S. Grassia, ``{Practical Parameterization of Rotations using the Exponential
  Map},'' \emph{J. Graphics Tools}, vol.~3, no.~3, pp. 29--48, 1998.

\bibitem{HS02}
E.~Hameiri and I.~Shimshoni, ``{Estimating the Principal Curvatures and the
  Darboux Frame from Real 3D Range Data},'' in \emph{3DPVT}, 2002, pp.
  258--267.

\bibitem{RVC02}
L.~Rocha, L.~Velho, and P.~C.~P. Carvalho, ``Image moments-based structuring
  and tracking of objects,'' in \emph{Brazilian Symposium on Computer Graphics
  and Image Processing}, 2002.

\bibitem{Ribeiro04}
M.~I. Ribeiro, ``{Gaussian Probability Density Functions: Properties and Error
  Characterization},'' Tech. Rep., 2004.

\bibitem{PKPSTV93}
J.~Ponce \emph{et~al.}, ``{Representations and Algorithms for 3D Curved Object
  Recognition},'' in \emph{Three-Dimensional Object Recognition Systems}.\hskip
  1em plus 0.5em minus 0.4em\relax Elsevier Press, 1993, pp. 327--352.

\bibitem{Eberly1999}
D.~Eberly, ``{Distance from Point to a General Quadratic Curve or a General
  Quadric Surface},'' Tech. Rep., 1999.

\bibitem{Taubin91}
G.~Taubin, ``{Estimation of Planar Curves, Surfaces, and Nonplanar Apace Curves
  Defined by Implicit Equations with Applications to Edge and Range Image
  Segmentation},'' \emph{IEEE Tran.on Pat. Anal. and Mach.Intel.,}, vol.~13,
  no.~11, pp. 1115--1138, 1991.

\bibitem{PF06}
S.~Paris and F.~Durand, ``{A Fast Approximation of the Bilateral Filter using a
  Signal Processing Approach},'' in \emph{In Proceedings of the European
  Conference on Computer Vision}, 2006, pp. 568--580.

\bibitem{LVJ05}
C.~H. Lee, A.~Varshney, and D.~W. Jacobs, ``{Mesh Saliency},'' in \emph{ACM
  SIGGRAPH}.\hskip 1em plus 0.5em minus 0.4em\relax New York, NY, USA: ACM,
  2005, pp. 659--666.

\bibitem{LG12}
J.~Lam and M.~Greenspan, ``{On the Repeatability of 3D Point Cloud Segmentation
  Based on Interest Points},'' in \emph{Computer and Robot Vision (CRV), 2012
  Ninth Conference on}, 2012, pp. 9--16.

\bibitem{Marigold08}
D.~S. Marigold, ``{Role Of Peripheral Visual Cues In Online Visual Guidance Of
  Locomotion},'' \emph{Exerc. Sport Sci. Rev.}, pp. 145--151, 2008.

\bibitem{HRDGN12}
S.~Holzer, R.~B. Rusu, M.~Dixon, S.~Gedikli, and N.~Navab, ``{Adaptive
  Neighborhood Selection for Real-Time Surface Normal Estimation from Organized
  Point Cloud Data Using Integral Images},'' in \emph{IEEE/RSJ Int. Conf. on
  Intel. Rob. and Syst. (IROS)}, 2012, pp. 2684--2689.

\bibitem{Bentley75}
J.~L. Bentley, ``{Multidimensional Binary Search Trees Used for Associative
  Searching},'' \emph{Comm. of the ACM}, vol.~18, no.~9, pp. 509--517, 1975.

\bibitem{SC86}
R.~Smith and P.~Cheeseman, ``{On the Representation and Estimation of Spatial
  Uncertainty},'' \emph{The International Journal of Robotics Research},
  vol.~5, pp. 56--68, 1986.

\bibitem{DB06}
H.~Durrant-Whyte and T.~Bailey, ``{Simultaneous Localization and Mapping: Part
  I,II},'' \emph{IEEE Rob. Autom. Mag.}, vol.~13, no.~2, pp. 99--110, 2006.

\bibitem{IKHMNKSHFDF11}
S.~Izadi \emph{et~al.}, ``{KinectFusion: Real-time 3D Reconstruction and
  Interaction Using a Moving Depth Camera},'' in \emph{24th Annual ACM Symp. on
  User Interf. Soft. and Tech.}, 2011, pp. 559--568.

\bibitem{CV96}
B.~Curless and M.~Levoy, ``{A Volumetric Method for Building Complex Models
  from Range Images},'' in \emph{23rd Annual Conf. on Comp. Graph. and Int.
  Tech.}, 1996, pp. 303--312.

\bibitem{PSLHS98}
S.~Parker, P.~Shirley, Y.~Livnat, C.~Hansen, and P.-P. Sloan, ``{Interactive
  Ray Tracing for Isosurface Rendering},'' in \emph{Conference on
  Visualization}, 1998, pp. 233--238.

\bibitem{LC87}
W.~E. Lorensen and H.~E. Cline, ``{Marching Cubes: A High Resolution 3D Surface
  Construction Algorithm},'' in \emph{14th Annual Conf. on Comp. Graph. and
  Int. Tech. (SIGGRAPH)}, 1987, pp. 163--169.

\bibitem{SHT09}
A.~Segal, D.~Haehnel, and S.~Thrun, ``{Generalized-ICP},'' in \emph{Proceedings
  of Robotics: Science and Systems}, Seattle, USA, June 2009.

\bibitem{HTA11}
I.~Ha, Y.~Tamura, H.~Asama, J.~Han, and D.~Hong, ``{Development of Open
  Humanoid Platform DARwIn-OP},'' in \emph{SICE Annual Conference}, 2011, pp.
  2178--2181.

\bibitem{Kanoulas14}
D.~Kanoulas, ``{Curved Surface Patches for Rough Terrain Perception},'' Ph.D.
  dissertation, College of Computer and Information Science, Northeastern
  University, 2014.

\bibitem{KZNKCT2017}
D.~Kanoulas, C.~Zhou, A.~Nguyen, G.~Kanoulas, D.~G. Caldwell, and N.~G.
  Tsagarakis, ``{Vision-Based Foothold Contact Reasoning using Curved Surface
  Patches},'' in \emph{IEEE Int. Conf. on Humanoid Rob. (Humanoids)}, 2017.

\bibitem{KWRZTSCT2018}
D.~Kanoulas, A.~Stumpf, V.~S. Raghavan, C.~Zhou, A.~Toumpa, D.~G.~C. Oskar~von
  Stryk, and N.~G. Tsagarakis, ``{Footstep Planning in Rough Terrain for
  Bipedal Robots using Curved Contact Patches},'' in \emph{IEEE International
  Conference on Robotics and Automation (ICRA)}, 2018.

\bibitem{KLCT2016}
D.~Kanoulas, J.~Lee, D.~G. Caldwell, and N.~G. Tsagarakis, ``{Visual Grasp
  Affordance Localization in Point Clouds Using Curved Contact Patches},''
  \emph{International Journal of Humanoid Robotics}, vol.~14, no.~01, p.
  1650028, 2016.

\bibitem{KTV2016}
D.~Kanoulas, N.~G. Tsagarakis, and M.~Vona, ``{Uncertainty Analysis for Curved
  Surface Contact Patches},'' in \emph{IEEE-RAS 16th Int. Conf. on Humanoid
  Robots (Humanoids)}, 2016, pp. 359--365.

\end{thebibliography}

\end{document}